\documentclass{article}

\usepackage{PRIMEarxiv}

\usepackage[utf8]{inputenc} 
\usepackage[T1]{fontenc}    
\usepackage{hyperref}       
\usepackage{url}            
\usepackage{booktabs}       
\usepackage{amsfonts}       
\usepackage{nicefrac}       
\usepackage{microtype}      
\usepackage{lipsum}
\usepackage{fancyhdr}       
\usepackage{graphicx}       
\usepackage{caption,subcaption}
\usepackage{amssymb}
\usepackage{amsmath}
\graphicspath{{media/}}     

\title{Generalizable End-to-End Deep Learning Frameworks for Real-Time Attitude Estimation Using 6DoF Inertial Measurement Units
}

\author{
  Arman Asgharpoor Golroudbari \\
  Department of Aerospace, \\Faculty of New Sciences \& Technologies, \\University of Tehran,\\
  Tehran, Iran \\
  a.asgharpoor@ut.ac.ir
  \And
  Mohammad Hossein Sabour \\
  Department of Aerospace, \\Faculty of New Sciences \& Technologies, \\ University of Tehran, \\
  Tehran, Iran \\
  sabourmh@ut.ac.ir
}

\begin{document}
\maketitle

\begin{abstract}
This paper presents a novel end-to-end deep learning framework for real-time inertial attitude estimation using 6DoF IMU measurements. Inertial Measurement Units are widely used in various applications, including engineering and medical sciences. However, traditional filters used for attitude estimation suffer from poor generalization over different motion patterns and environmental disturbances. To address this problem, we propose two deep learning models that incorporate accelerometer and gyroscope readings as inputs. These models are designed to be generalized to different motion patterns, sampling rates, and environmental disturbances. Our models consist of convolutional neural network layers combined with Bi-Directional Long-Short Term Memory followed by a Fully Forward Neural Network to estimate the quaternion. We evaluate the proposed method on seven publicly available datasets, totaling more than 120 hours and 200 kilometers of IMU measurements. Our results show that the proposed method outperforms state-of-the-art methods in terms of accuracy and robustness. Additionally, our framework demonstrates superior generalization over various motion characteristics and sensor sampling rates. Overall, this paper provides a comprehensive and reliable solution for real-time inertial attitude estimation using 6DoF IMUs, which has significant implications for a wide range of applications.

\end{abstract}


\section{Introduction}
Attitude estimation is a fundamental problem in robotics, navigation, and aerospace engineering. It refers to the task of determining the orientation of an object relative to a reference coordinate system. This information is crucial for many applications, such as controlling the orientation of a spacecraft or a quadcopter, stabilizing a camera or a sensor platform, and enabling precise navigation in GPS-denied environments \cite{al2019deep}.

Inertial sensors, such as accelerometers and gyroscopes, are commonly used for attitude estimation due to their low cost, small size, and high accuracy. However, estimating attitude from raw sensor measurements is a challenging problem due to the nonlinearities and biases in the sensor data, as well as the presence of external disturbances and measurement noise.

Over the past few decades, various approaches have been proposed for attitude estimation, ranging from traditional linear and nonlinear filters to more recent machine learning-based methods. In particular, deep learning has emerged as a promising technique for attitude estimation, as it can learn complex mappings from raw sensor data to attitude estimates without relying on handcrafted features or prior knowledge of the system dynamics.

Numerous instruments and sensors are available for this purpose, but they vary in cost and complexity. While high-quality sensors can provide more accurate results, they may not always be practical due to their high cost. One way to increase accuracy at a lower cost is to use multiple sensors, either of the same type (homogenous) or different types (heterogeneous). This approach, known as Multi-Data Sensor Fusion (MSDF). MDSF is a technique that combines information from multiple sensors of different types or modalities to improve the accuracy and reliability of a system's perception or estimation. In other words, MDSF integrates data from multiple sensors that capture different aspects of the environment or the system being monitored in order to obtain a more complete and robust understanding of the underlying phenomena. MSDF can be further divided into two categories: single-point methods, which use vector measurements at a single point in time, and recursive methods, which combine measurements over time and the system's mathematical model \cite{gebre2004design}.

The precision of attitude determination depends on the sensors' accuracy, the system modeling quality, and the information processing method \cite{renaudin2014magnetic}. Achieving this level of precision is a challenging navigation problem due to system modeling, process, and measurement errors. Increasing the sensor's precision may exponentially increase the cost, and sometimes, achieving the precision requirements is only possible for an exorbitant cost.

Inertial navigation algorithms, which are based on the Dead Reckoning method, have been used for years to determine the attitude based on inertial sensors \cite{steinhoff2010dead}. The method utilizes different types of inertial sensors such as accelerometers and gyroscopes, referred to as Inertial Measurement Units (IMUs) \cite{vertzberger2022adaptive}. A moving object's position, velocity, and attitude can be determined using the numerical integration of IMU measurements. Over the past decade, MEMS-based IMUs have become increasingly popular. Due to recent advances in MEMS technology, IMUs have become smaller, cheaper, and more accurate. They can now be found in mobile robots, smartphones, drones, and autonomous vehicles. However, IMUs suffer from noise and bias, which directly affect the performance of the attitude estimation algorithm \cite{mahdi2022machine}.

Real-time attitude estimation based on IMU sensor raw data is a fundamental problem in sensor fusion. In the past decades, various MSDF techniques and Deep Learning models have been developed to tackle this problem and increase the accuracy and reliability of attitude estimation techniques. Attitude can be estimated using at least a 6-Degree-of-Freedom (6DoF) Sensor Fusion Algorithm (SFA) in MSDF methods. In 6DoF SFAs, a three-axis accelerometer is fused with a three-axis gyroscope to estimate the attitude. However, 6DoF SFAs are not suitable for attitude and heading estimation/determination as the accelerometer cannot measure the yaw (heading) angle, and the gyroscope can only measure the yaw angle's rate \cite{ding2022improved}. An alternative method is to fuse magnetometer readings with a 6DoF SFA to estimate the full orientation (attitude and heading). However, a magnetometer's primary disadvantage is the magnetic disturbances, which adversely affect its performance, mainly when used for indoor navigation. Several techniques have been developed to reduce the effect of magnetic disturbances on the filter performance, such as the Factorized Quaternion Algorithm (FQA) \cite{lee2012factorized}.

Most SFAs are developed and parametrized based on the system's dynamic model, which requires a precise choice of model parameters \cite{fauske2007estimation}. To the best of our knowledge, no algorithm can handle all types of motions, and it is challenging to design a generalizable algorithm for all possible scenarios. The traditional approach is to develop a model for a specific scenario and then adapt it to different situations. However, this approach is time-consuming and requires extensive domain knowledge.

Recently, deep learning models have shown great potential in solving sequential data problems, including attitude estimation \cite{hoang2022yaw,asgharpoor2022design}, error modeling \cite{zhao2022attitude}. These models can learn the hidden patterns and relationships within the data and handle complex and nonlinear relationships between sensor measurements and attitude. Several studies have shown the potential of deep learning models for attitude estimation, including recurrent neural networks (RNNs) and convolutional neural networks \cite{wang2020recent,xiao2018opportunities,zulqarnain2020comparative,nevavuori2020crop,bouktif2019single}.

The field of robotic perception has seen significant research attention in recent years, with many studies focusing on the development of odometry or the fusion of heterogeneous sensors for attitude estimation, including inertial-GPS or inertial-visual fusion, and learning-based frameworks coupled with conventional filtering methods. However, there remains a notable gap in research efforts devoted to end-to-end learning-based inertial attitude estimation. Although relying solely on inertial data can result in significant drift or biases, the benefits of developing such models in scenarios where visual data is not available or GPS is denied cannot be understated. Therefore, there is a pressing need to design a learning-based model that can achieve end-to-end inertial attitude estimation, thereby providing highly accurate and reliable estimates of a system's orientation based solely on inertial sensor data. Such a model holds great potential for a range of applications, including robotics, navigation, and aerospace engineering.

To the best of our knowledge, there are currently only three models that have focused on utilizing end-to-end learning-based methods for inertial attitude estimation. The first model, RIANN \cite{weber2021riann}, introduced a GRU-based approach. While the trained model is publicly available, their article lacked sufficient quantitative information, making it difficult to reproduce their results. In another study \cite{narkhede2021incremental}, the authors presented an LSTM-based model, but their work was limited to only one sampling rate, which was not mentioned in their publication, and was not tested on publicly available inertial datasets. Lastly, the third study \cite{brotchie2022leveraging} employed a self-attention mechanism. However, their model was trained and tested only on the OxIOD dataset \cite{chen2018oxiod}, which makes it limited to only one type of motion and a specific sampling rate.  The objective of this study is to introduce a novel end-to-end learning framework for estimating orientation, independent of external data sources and generalize across various environments and sensors sampling rate, and thereby contribute to this critical area of research.

This study presents three novel end-to-end deep learning architectures for real-time attitude estimation, which are implemented using low-cost strapdown inertial measurement units based on micro-electromechanical systems (MEMS). Our proposed approaches utilize both RNNs and a hybrid RNN-CNN neural network, which can effectively learn and model the motion characteristics, noise, and bias associated with inertial sensor measurements across various devices. Our models estimate the roll and pitch angles from raw sensor measurements and compute the quaternion using a fully connected neural network. Specifically, the models receive the IMU readings in a window of 100 frames containing 3-axis angular velocity and 3-axis acceleration as inputs. To evaluate the performance of our models, we conducted both qualitative and quantitative assessments using publicly available inertial datasets. The results demonstrate that our proposed quaternion multiplicative error-based loss functions outperformed recent inertial estimation methods by up to 40\%, producing the most accurate inertial estimation results. Our study highlights the efficacy of end-to-end deep learning approaches for real-time attitude estimation in low-cost IMUs, and their potential for various applications, including drones and autonomous vehicles.

Our comprehensive analysis of the proposed approach using various publicly available IMU datasets has demonstrated superior performance in terms of accuracy and robustness compared to traditional algorithms and other deep learning model. Our main contribution in this research is the presentation of a novel end-to-end deep learning framework for inertial attitude estimation in different scenarios, such as indoor and outdoor navigation, drones, and autonomous vehicles.

The paper is structured as follows: Section 2 presents a review of related studies on attitude estimation using deep learning models. Section 3 provides a detailed description of the problem, including system modeling, sensor measurements, and error sources. In Section 4, we describe the proposed approach, including the network architecture, loss function, and training procedure. Section 5 presents a comprehensive description of our experiments, including the datasets used, evaluation metrics, and comparison with state-of-the-art approaches. The results and analysis of the tests are presented in Section 6. Finally, we draw conclusions and outline future research directions in Section 7.

\section{Literature Review}

A 3D dead-reckoning navigation system, such as an Inertial Navigation System (INS), includes a set of Inertial Measurement Units that consist of three gyroscopes and three mutually orthogonal accelerometers. This system also incorporates a navigation processor that integrates the IMU's outputs to provide information about position, velocity, and attitude \cite{groves2015principles}. While these navigation systems are widely used in various fields, including medical science and aerospace, they suffer from a significant amount of noise and bias in their measurements. This accumulation of errors over time makes them unsuitable for long-term use.

In the past decade, various research efforts have been made to address this problem by improving inertial navigation techniques. These efforts can be classified into three categories: estimation methods, MSDF techniques, and evolutionary/AI algorithms. Estimation methods, such as the Kalman Filter (KF) family (i.e., EKF, UKF, MEKF), and other commonly used algorithms, such as Madgwick \cite{madgwick2010efficient} and Mahony \cite{euston2008complementary}, are based on the dynamic model of the system. The Kalman Filter was initially introduced in \cite{kalman1960}, and its variants, including EKF \cite{jing2017attitude}, UKF \cite{chiella2019quaternion}, and MEKF \cite{hall2008quaternion}, have been implemented for attitude estimation applications \cite{crassidis2007survey}. In a recent comparative study, Caruso et al. \cite{caruso2021analysis} compared various sensor fusion algorithms for inertial attitude estimation and demonstrated that SFA performance is highly dependent on parameter tuning. Fixed parameter values are unsuitable for all applications, and parameter tuning is one of the drawbacks of conventional attitude estimation methods.

Evolutionary algorithms, such as fuzzy logic \cite{shen2012adaptive,widodo2014complementary}, and deep learning \cite{brossard2020denoising,han2021hybrid,buchanan2022deep,engelsman2022data}, could overcome this problem. Most deep learning methods in inertial navigation have focused on inertial odometry \cite{esfahani2019aboldeepio,aslan2022hvionet,soyer2021efficient,saha2022tinyodom,onyekpe2021io,guimaraes2021deep,lin2022residual}, while only a few have attempted to solve the inertial attitude estimation problem \cite{weber2021riann,esfahani2019orinet}. Deep learning methods are typically used for visual or visual-inertial-based navigation \cite{aslan2022visual,ozaki2021dnn,yu2019hybrid,fan2021fast}.

In recent years, various studies have explored the application of neural networks to satellite attitude estimation. One such study by Rochefort et al. \cite{rochefort2005new} introduced a novel approach that employs a quaternion neural network for state estimation. This method is characterized by high accuracy and significantly lower computational complexity when compared to the traditional Extended Kalman Filter (EKF).

Another approach proposed by Chang et al. \cite{chang2011time} involves the use of a Time-Varying Complementary Filter (TVCF) in combination with a fuzzy logic inference system to adjust the Complementary Filter (CF) parameters for attitude estimation. Chen et al. \cite{chen2018ionet,rochefort2005new} utilized deep recurrent neural networks to estimate user displacement over a specified time interval.

Esfahani et al. \cite{esfahani2019orinet} introduced OriNet, a method that estimates orientation in quaternion form based on Long Short-Term Memory (LSTM) layers and IMU measurements. Zhang et al. \cite{zhang2019fusion} developed a sensor fusion method that utilizes empirical mode decomposition threshold filtering (EMDTF) to eliminate IMU noise and a LSTM neural network to predict pseudo-GPS position during GPS outages.

Dhahbane et al. \cite{dhahbane2020neural} proposed a neural network-based Complementary Filter (NNCF) with ten hidden layers that is trained by the Bayesian Regularization Backpropagation (BRB) training algorithm to improve generalization qualities and solve the overfitting problem. Li et al. \cite{li2019novel} suggested an adaptive Kalman filter with a fuzzy neural network for trajectory estimation system. This algorithm helps to mitigate measurement noise and undulation for implementing the touch interface. Finally, Brossard et al. \cite{brossard2020denoising} employed deep learning to denoise gyroscope measurements for an open-loop attitude estimation algorithm.

Weber et al. \cite{weber2021riann} have presented a real-time capable neural network for robust attitude estimation based on an IMU. Their model takes input from an accelerometer, gyroscope, and IMU sampling rate and outputs the attitude in quaternion form. However, it is only capable of estimating the roll and pitch angle. In \cite{sun2021idol}, Sun et al. introduced a two-stage deep learning framework for inertial odometry based on a Long-Short-Term Memory and Feed-Forward Neural Network (FFNN) architecture. The first stage estimates the orientation, while the second stage estimates the position.

Santos et al. \cite{dos2021static} developed a Neural Network model for static attitude determination based on the PointNet architecture. They used an attitude profile matrix as input, and the Swish activation function and Adam optimizer were employed. Al et al. \cite{al2019deep} developed a deep learning model for estimating the Multirotor Unmanned Aerial Vehicle (MUAV) based on the Kalman filter and FFNN. In \cite{narkhede2021incremental}, Narkhede et al. utilized the LSTM framework to estimate Euler angles using an accelerometer, gyroscope, and magnetometer, but they did not consider the sensor sampling rate.

In Table~\ref{tab1}, we have summarized some related works in the navigation field that have utilized deep learning techniques.

\begin{table*}[!ht]
\caption{Deep Learning for Localization.\label{tab1}}
\centering
\begin{tabular}{lccc}
  \hline
\textbf{Model} & \textbf{Year/Month} & \textbf{Input Data} & \textbf{Application}\\
\hline
PoseNet\cite{kendall2015posenet}          & 2015/12             & Vision                          & Relocalization            \\
VINet\cite{clark2017vinet}                & 2017/02             & Vision + Inertial               & Visual Inertial Odometry  \\
DeepVO\cite{wang2017deepvo}               & 2017/05             & Vision                          & Visual Odometry           \\
VidLoc\cite{clark2017vidloc}              & 2017/07             & Vision                          & Relocalization            \\
IONet\cite{chen2018ionet}                 & 2018/02             & Inertial                    & Inertial Odometry         \\
UnDeepVO\cite{li2018undeepvo}             & 2018/05             & Vision                          & Visual Odometry           \\
VLocNet\cite{valada2018deep}              & 2018/05             & Vision                          & Relocalization, Odometry  \\
RIDI\cite{yan2018ridi}                    & 2018/09             & Inertial                   & Inertial Odometry         \\
SIDA\cite{chen2020learning}               & 2019/01             & Inertial                    & Domain Adaptation         \\
VIOLearner\cite{shamwell2019unsupervised} & 2019/04             & Vision + Inertial               & Visual Inertial Odometry  \\
RINS-W\cite{brossard2019rins}             & 2019/05             & Inertial Only                   & Inertial Odometry         \\
SelectFusion\cite{chen2019selectfusion}   & 2019/06             & Vision + Inertial + LIDAR       & Visual Inertial Odometry and sensor Fusion     \\
LO-Net\cite{li2019net}                    & 2019/06             & LIDAR                           & LIDAR Odometry            \\
L3-Net\cite{lu2019l3}                     & 2019/06             & LIDAR                           & LIDAR Odometry            \\
Lima et al.\cite{silva2019end}            & 2019/8              & Inertial                        & Inertial Odometry         \\
DeepVIO\cite{han2019deepvio}              & 2019/11             & Vision + Inertial                 & Visual Inertial Odometry  \\
OriNet\cite{esfahani2019orinet}           & 2020/4              & Inertial                        & Inertial Odometry         \\
Sorg \cite{sorg2020thesis}                     & 2020/4              & Inertial                        & Pose Estimation           \\
GALNet\cite{mendoza2020galnet}            & 2020/5              & Inertial + Dynamic + Kinematic & Autonomous Cars           \\
PDRNet\cite{asraf2021pdrnet}              & 2021/3              & Inertial                        & Pedestrian Dead Reckoning \\
Kim et al.\cite{kim2021nine}              & 2021/4              & Inertial                        & Inertial Odometry         \\
RIANN\cite{weber2021riann}                & 2021/5              & Inertial                        & Attitude Estimation       \\
CTIN \cite{rao2022ctin}                   & 2022/6              & Inertial                        & Inertial Odometry         \\
Xia et al. \cite{xia2022faster}           & 2022/8              & Inertial                        & Human Pose Estimation     \\
Brotchie et al. \cite{brotchie2022leveraging}& 2022/11              & Inertial                        & Attitude Estimation     \\
\hline  
\end{tabular}
\end{table*}

Most studies have primarily focused on odometry or coupling learning-based methods with traditional filtering algorithms. Consequently, there is a significant gap in research regarding end-to-end learning-based inertial attitude estimation. As mentioned in the introduction section, only three end-to-end deep learning frameworks have been proposed, with RIANN being the only one that has been demonstrated to generalize across various sampling rates.

\section{Problem definition}

The objective of this article is to tackle the challenge of real-time estimation of an object's attitude using an IMU sensor and generalized across different sampling rates. The primary obstacle to this is the significant level of noise and bias present in the measurements, leading to error accumulation over time and decreased accuracy of the attitude estimate. The aim is to develop a reliable and accurate method for estimating the object's attitude in real-time, which does not require an initial reset period for filter convergence. This is essential for applications such as navigation, image stabilization, tracking, and autonomous vehicles, where precise attitude determination is crucial for successful performance.

Attitude determination and control are critical aspects of the field of Aerospace Engineering. One of the essential components of this field is pointing modes such as Earth pointing or Sun pointing, which requires subsystems to be directed in specific directions. Accurate knowledge of the vehicle's attitude, i.e., its orientation relative to a reference frame, is necessary to achieve proper orientation. Attitude determination methods are categorized as either static or dynamic.

Static attitude determination is a time-independent approach that relies on measurements or observations to obtain information describing the object's orientation relative to a reference frame. This method involves measuring the directions from the vehicle to known points, commonly referred to as Attitude Knowledge. However, deterministic approaches often fall short in accuracy due to measurement noise, model error, and process error. Therefore, statistical methods can be used to provide better accuracy.

Dynamic attitude determination, also known as attitude estimation, uses mathematical techniques such as statistical and probabilistic methods to predict and estimate the future attitude based on a dynamic model and prior measurements. These methods employ data fusion techniques, which integrate a series of measurements using filtering algorithms or Multi-Sensor-Data-Fusion. The most commonly used attitude estimation methods are the Extended Kalman Filter, Madgwick, and Mahony. These methods improve attitude determination accuracy and enable better pointing control.

In this study, we propose a novel approach based on a Neural Network model that takes current and previous gyroscope and accelerometer measurements as input and estimates the attitude. Our approach does not consider any initial reset period for filter convergence, making it more efficient and suitable for real-time applications compared to previous studies.In this article, we aim to address the problem of real-time estimation of an object's attitude using an IMU sensor, which measures its angular velocity and linear acceleration using gyroscopes and accelerometers. The primary challenge of this problem is the significant level of noise and bias present in the measurements, leading to error accumulation over time and decreased accuracy of the attitude estimate. Our goal is to develop a reliable and accurate method for estimating the object's attitude in real-time without requiring an initial reset period for filter convergence. This is crucial for various applications, including navigation, image stabilization, tracking, and autonomous vehicles, where precise attitude determination is vital for successful performance. We propose a novel approach based on a Neural Network model that takes current and previous gyroscope and accelerometer measurements as input and estimates the attitude. Unlike previous studies, our approach does not consider any initial reset period for filter convergence, making it more efficient and suitable for real-time applications.

\section{Methodology}
\subsection{Error Matrices}
This section explores the significance of error matrices in the field of aerospace engineering for precise estimation of a vehicle's orientation using quaternions. The purpose of error matrices is to quantify the accuracy of the estimation algorithm by establishing a connection between the error in the estimated attitude and the error in the measured sensor data. The section highlights the need for selecting appropriate error matrices for specific applications and desired properties of the estimated attitude and the importance of careful tuning of these parameters to achieve optimal results.

In addition, the section emphasizes the importance of choosing an appropriate loss function in deep neural network training to measure the difference between predicted and actual output values. Since the attitude error is a geometric quantity, conventional algebraic error matrices such as mean squared error or mean absolute error are unsuitable for this purpose. Therefore, various methods such as geodesic distances have been developed to define and implement quaternion error. The section outlines the most commonly used quaternion error measures for attitude estimation, including quaternion rotational error, quaternion inner products, and error matrices. The section underscores the importance of careful selection and implementation of error matrices to ensure accurate and reliable attitude estimation.

In conclusion, error matrices are vital for accurate attitude estimation, which is critical for aerospace engineering applications. They play a crucial role in defining the loss function that guides the optimization process and measures the model's performance in deep learning. The choice of error representation method, particularly in the context of attitude estimation, should be carefully considered to ensure accuracy and reliability. The following section will introduce the most widely used quaternion error measures, including the quaternion inner product, which can be used to train deep learning models.

Accurate estimation of a vehicle's orientation is critical in aerospace engineering. One effective method for estimating orientation is through the use of quaternions, which represent rotation around a unit vector \cite{bani2012survey}. In the context of attitude estimation, error matrices play a crucial role in quantifying the accuracy of an estimation algorithm. These matrices relate the error in the estimated attitude to the error in the measured sensor data. There are various types of error matrices that can be used, and the selection depends on the specific application and desired properties of the estimated attitude. It is important to carefully choose and tune these parameters, including the training data and network architecture, to achieve optimal results.

In deep learning, a loss function is used to measure the difference between predicted and actual output values. The loss function plays a critical role in the training process of a deep neural network, as it guides the optimization process and measures the model's performance. In the context of attitude estimation, the loss function measures the difference between the predicted attitude (e.g., quaternion or Euler angles) and the true attitude of the vehicle. The most commonly used loss functions include mean squared error (MSE) and mean absolute error (MAE).

However, as the attitude error is a geometric quantity, algebraic error matrices such as the MSE or MAE are not suitable. Instead, several methods have been developed for defining and implementing quaternion error. These methods include the use of geodesic distances and Riemannian metrics. Careful consideration should be given to the choice of the error representation method in order to ensure accurate and reliable attitude estimation.

In the following, we will introduce the most widely used quaternion error measures for attitude estimation that can be employed for training deep learning models.

A quaternion can be expressed as a vector consisting of the scalar component $q_w$ and the vector components $q_x$, $q_y$, and $q_z$ as shown below:

\begin{equation}
\begin{gathered}
q =
\begin{bmatrix}
\cos(\theta / 2) \\
\sin(\theta / 2) \cdot u_x \\
\sin(\theta / 2) \cdot u_y \\
\sin(\theta / 2) \cdot u_z
\end{bmatrix}
=
\begin{bmatrix}
q_w \\
q_x \\
q_y \\
q_z
\end{bmatrix}
\end{gathered}
\end{equation}

where $\theta$ is the angle of rotation and $\hat{u}$ is the unit vector of the rotation axis.
Attitude also could be defined as a rotation from the true orientation to the estimated orientation. 
\begin{equation}
	\begin{gathered}
		\mathbf{q}_{est} = \delta \mathbf{q} \otimes \mathbf{q}_{true}
	\end{gathered}
\end{equation}
where $\mathbf{q}_{est}$ is the estimated quaternion, $\mathbf{q}_{true}$ is the true quaternion, and $\delta \mathbf{q}$ is the quaternion error. 

The attitude error or the quaternion rotational error can be expressed as the difference between the estimated quaternion $\mathbf{q}_{est}$ and the true quaternion $\mathbf{q}_{true}$:

\begin{equation}
\begin{gathered}
\mathbf{q}_{err} = \mathbf{q}_{true} \otimes \mathbf{q}_{est}^{-1}
\end{gathered}
\end{equation}

The error rate can be calculated using the following equation:

\begin{equation}
\begin{gathered}
\dot{\mathbf{q}}_{err} = \dot{\mathbf{q}}_{true} \otimes \mathbf{q}_{est}^{-1} + \mathbf{q}_{true} \otimes \dot{\mathbf{q}}_{est}^{-1}
\end{gathered}
\end{equation}

A simpler method to calculate the error is to use element-wise subtraction between the true and estimated quaternions:

\begin{equation}
\begin{gathered}
\mathbf{q}_{err} = \mathbf{q}_{true} - \mathbf{q}_{est}
\end{gathered}
\end{equation}

Error matrices are used to represent the covariance between the errors of each element in the quaternion. The error matrices can be defined as:

\begin{equation}
\begin{gathered}
\mathbf{P}_{err} = E{(\mathbf{q}_{err} - \bar{\mathbf{q}}_{err})(\mathbf{q}_{err} - \bar{\mathbf{q}}_{err})^T}
\end{gathered}
\end{equation}

where $\bar{\mathbf{q}}_{err}$ is the mean quaternion error, and $E{\cdot}$ is the expected value operator.

The Quaternion Inner Product (QIP) is a measure of the angle between two quaternions, which represents the difference between the predicted and true orientation. The dot product between two quaternions is equivalent to the angle between two points on the quaternion hypersphere. The QIP is defined as follows:

\begin{equation}
\begin{gathered}
QIP(q,p) = q \cdot p = q_w p_w + q_x p_x + q_y p_y + q_z p_z
\end{gathered}
\end{equation}

The QIP yields the quaternion difference between two quaternions, so if the angle between two quaternions is zero, the QIP value will be equal to 1. Therefore, the QIP loss function can be defined as:

\begin{equation}
\begin{gathered}
L_{QIP} = \frac{1}{\mathcal{N}} \sum_{i=1}^{\mathcal{N}} (1 - | q \cdot p | )
\end{gathered}
\end{equation}

On the other hand, the angle between two quaternions can be computed using the quaternion inner product, as follows:

\begin{equation}
\begin{gathered}
L_{QIPA} = \frac{1}{\mathcal{N}} \sum_{i=1}^{\mathcal{N}} (\theta) = \frac{1}{\mathcal{N}} \sum_{i=1}^{\mathcal{N}} (\arccos(q \cdot p))
\end{gathered}
\end{equation}

In \cite{brotchie2022leveraging}, the authors utilized a combination of QIP and MSE, as follows:

\begin{equation}
\begin{gathered}
L_{QIP-MSE} = \frac{1}{\mathcal{N}} \sum_{i=1}^{\mathcal{N}} QIP({q_{true}^{i}-q_{est}^{i}}, {q_{est}^{i}-q_{true}^{i}})
\end{gathered}
\end{equation}

Here, $q_{true}^{i}$ and $q_{est}^{i}$ denote the true and estimated quaternions, respectively, and $\mathcal{N}$ is the number of samples in the dataset. The QIP-MSE loss function combines the geometric information of the QIP with the algebraic information of the MSE. This combination has been shown to produce more accurate and reliable results in certain applications.

The research conducted in \cite{silva2019end} employed the Quaternion Multiplicative Error (QME) loss function to assess the effectiveness of their proposed method utilizing the Hamilton product. The QME loss function was defined as follows:
\begin{equation}
\begin{gathered}
L_{QME} = \frac{1}{\mathcal{N}} \sum_{i=1}^{\mathcal{N}} \left(2 \cdot \lVert imag(q \otimes p^{\star}) \rVert 1\right)
\end{gathered}
\end{equation}
Here, $p^{\star}$ denotes the complex conjugate of quaternion $p$. The complex conjugate of a quaternion can be calculated by:
\begin{equation}
\begin{gathered}
p^{\star} = \begin{bmatrix}
p_0& -p_1 &-p_2 &-p_3
\end{bmatrix}^T
\end{gathered}
\end{equation}
Alternatively, the angle corresponding to the QME can be computed as follows:
\begin{equation}
\begin{gathered}
L{QMEA} = 2 \cdot \arccos(|\text{scalar}( q \otimes p^{\star}) |)
\end{gathered}
\end{equation}
However, the use of the $arccos$ function can lead to values greater than 1 or less than -1 for the scalar part of $(q \otimes p^{\star})$, which can cause the gradient to explode. To tackle this issue, the scalar part of $(q \otimes p^{\star})$ is clamped to the range of [-1, 1]. Although this value clipping could result in a loss of information about the angle between two quaternions, another approach to avoiding gradient explosion is replacing the $arccos$ function with a non-trigonometric linear function. The latter can be defined as follows:
\begin{equation}
\begin{gathered}
L_{QMEAnT} = 1 - \sqrt{(|\text{scalar}( q \otimes p^{\star}) |)^2}
\end{gathered}
\end{equation}
where $q_w$ and $q_z$ denote the squared values of $q \otimes p^{\star}$.
Similarly, the authors in \cite{laidig2021broad} decomposed the attitude error into a rotation around the $z$-axis, $e_h$, and the shortest residual rotation, $e_i$. These two angles can be computed using the following equations:
\begin{equation}
\begin{gathered}
e_h = 2 \arctan(q_z / q_w)
\end{gathered}
\end{equation}
\begin{equation}
\begin{gathered}
e_i = 2 \arccos ( \sqrt{q_w^2 + q_z^2})
\end{gathered}
\end{equation}
Furthermore, using the $arccos$ function in the computation of the $e_i$ angle could lead to instability in gradient calculation. As mentioned in \cite{weber2020neural}, the $arccos$ function can be replaced by:
\begin{equation}
\begin{gathered}
1 - \sqrt{q_w^2 + q_z^2}
\end{gathered}
\end{equation}

The loss function can be expressed as follows:

\begin{equation}
\begin{gathered}
L_{e_i} = \frac{1}{\mathcal{N}} \sum_{i=1}^{\mathcal{N}} \left(1 - \sqrt{q_w^2 + q_z^2}\right)
\end{gathered}
\end{equation}

Quaternion Shortest Geodesic Distance (QSGD) is the angle between the predicted and true orientation on the quaternion hyper-sphere. It is defined as:

\begin{equation}
\begin{gathered}
QSGD = q \otimes p^{\star} =
\begin{bmatrix}
q_w p_w - q_x p_x - q_y p_y - q_z p_z \\
q_w p_x + q_x p_w + q_y p_z - q_z p_y \\
q_w p_y - q_x p_z + q_y p_w + q_z p_x \\
q_w p_z + q_x p_y - q_y p_x + q_z p_w
\end{bmatrix}
\end{gathered}
\end{equation}

The loss function for QSGD is defined as follows:

\begin{equation}
\begin{gathered}
L_{QSGD} = | 1 - (|\text{scalar}( q \otimes p^{\star}) |) |
\end{gathered}
\end{equation}

Alternatively, the QSGD loss function can be expressed as:

\begin{equation}
\begin{gathered}
L_{QSGD2} = \sqrt{1-\sqrt{\text{scalar}( q \otimes p^{\star})^2}}
\end{gathered}
\end{equation}

By decomposing $q \otimes p^{\star}$ into its constituent parts, where $\hat{u}$ is the unit vector of the rotation axis, and $\theta$ is the rotation angle, it can be represented as:

\begin{equation}
\begin{gathered}
q \otimes p^{\star} = \begin{bmatrix} \cos(\theta_{err}/2) \\ \sin(\theta_{err}/2) \cdot \hat{u_{err}} \end{bmatrix} = \begin{bmatrix} w_{err} \\ x_{err} \\ y_{err} \\ z_{err} \end{bmatrix}
\end{gathered}
\end{equation}

When $\theta = 0$, the quaternion difference equals $q \otimes p^{\star} = \begin{bmatrix} 1 & 0 & 0 & 0 \end{bmatrix}^T$. The loss function that minimizes the rotation angle between two quaternions is defined as:

\begin{equation}
\begin{gathered}
L_{QSGD3} = \begin{bmatrix} w_{err}-1 \\ x_{err} \\ y_{err} \\ z_{err} \end{bmatrix}
\end{gathered}
\end{equation}

\begin{figure*}[!ht]
\centering
\includegraphics[width=.9\textwidth]{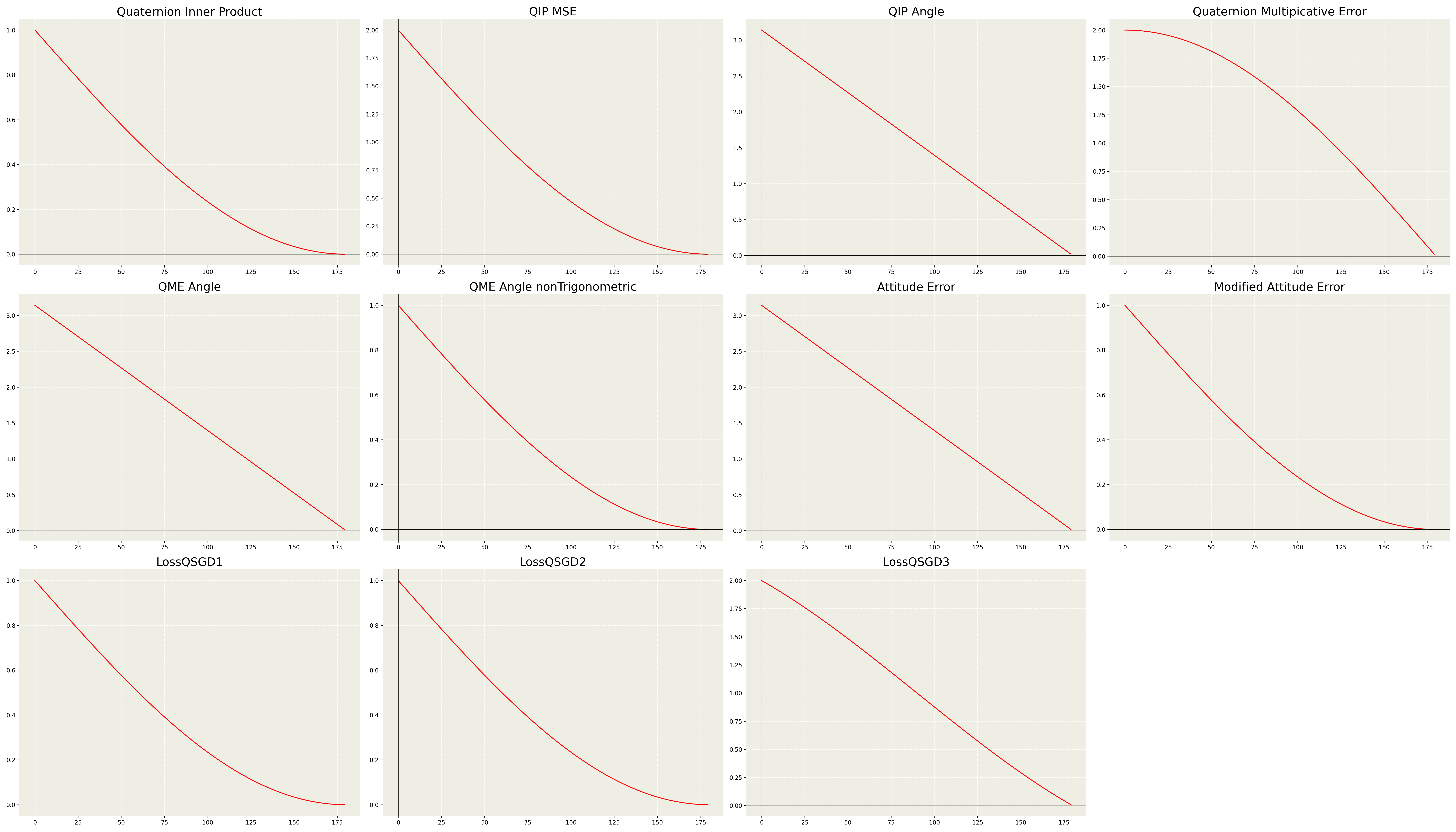}
\caption{Loss functions for attitude error}
\label{fig:loss_functions}
\end{figure*}

\begin{figure}[!ht]
\centering
  \centering
    \includegraphics[width=0.45\textwidth]{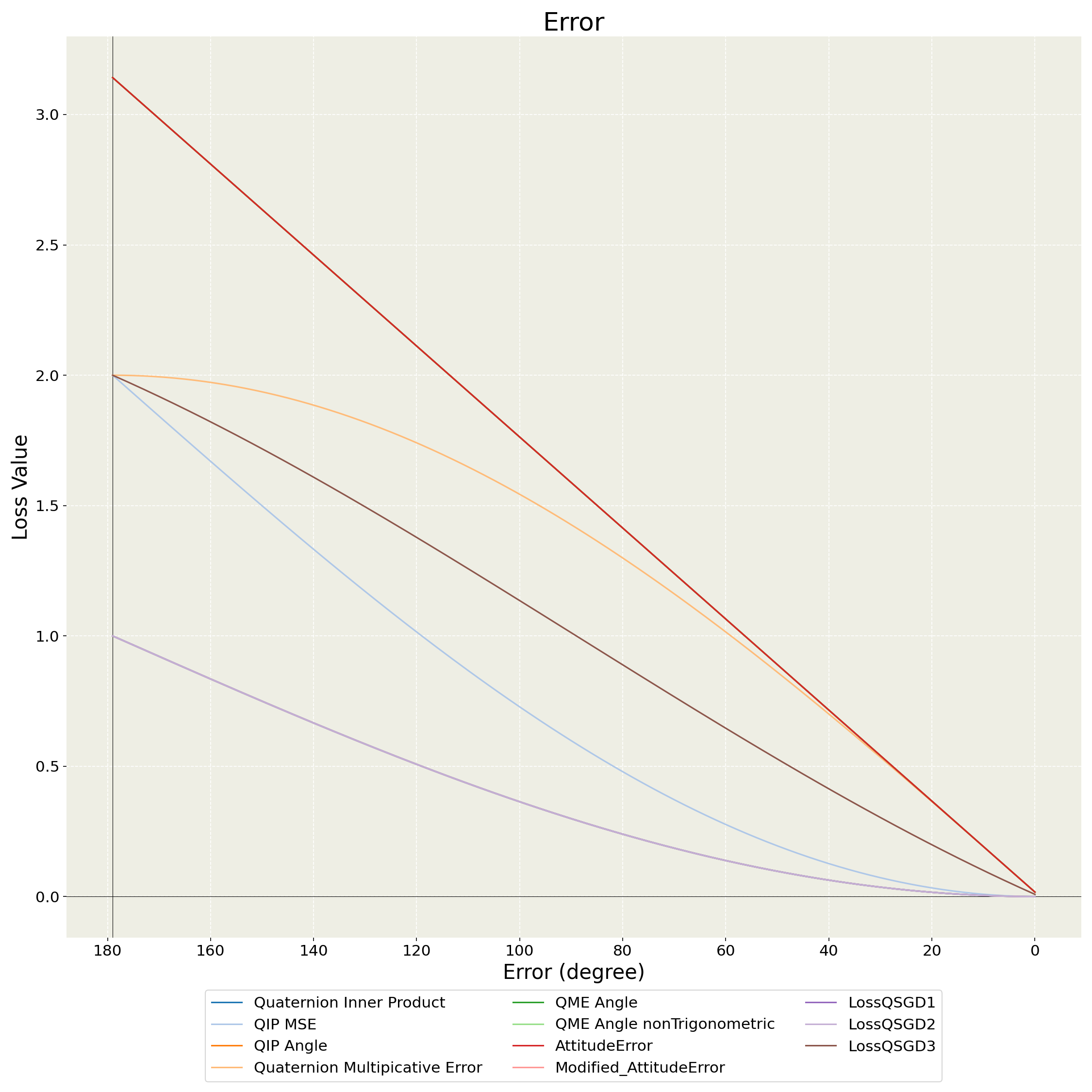}
    \caption{Compare Loss functions for attitude error \cite{golroudbari2019design}}
\label{fig:loss_functionsComp}%
\end{figure}

\begin{figure}[!ht]
    \centering
    \includegraphics[width=0.45\textwidth]{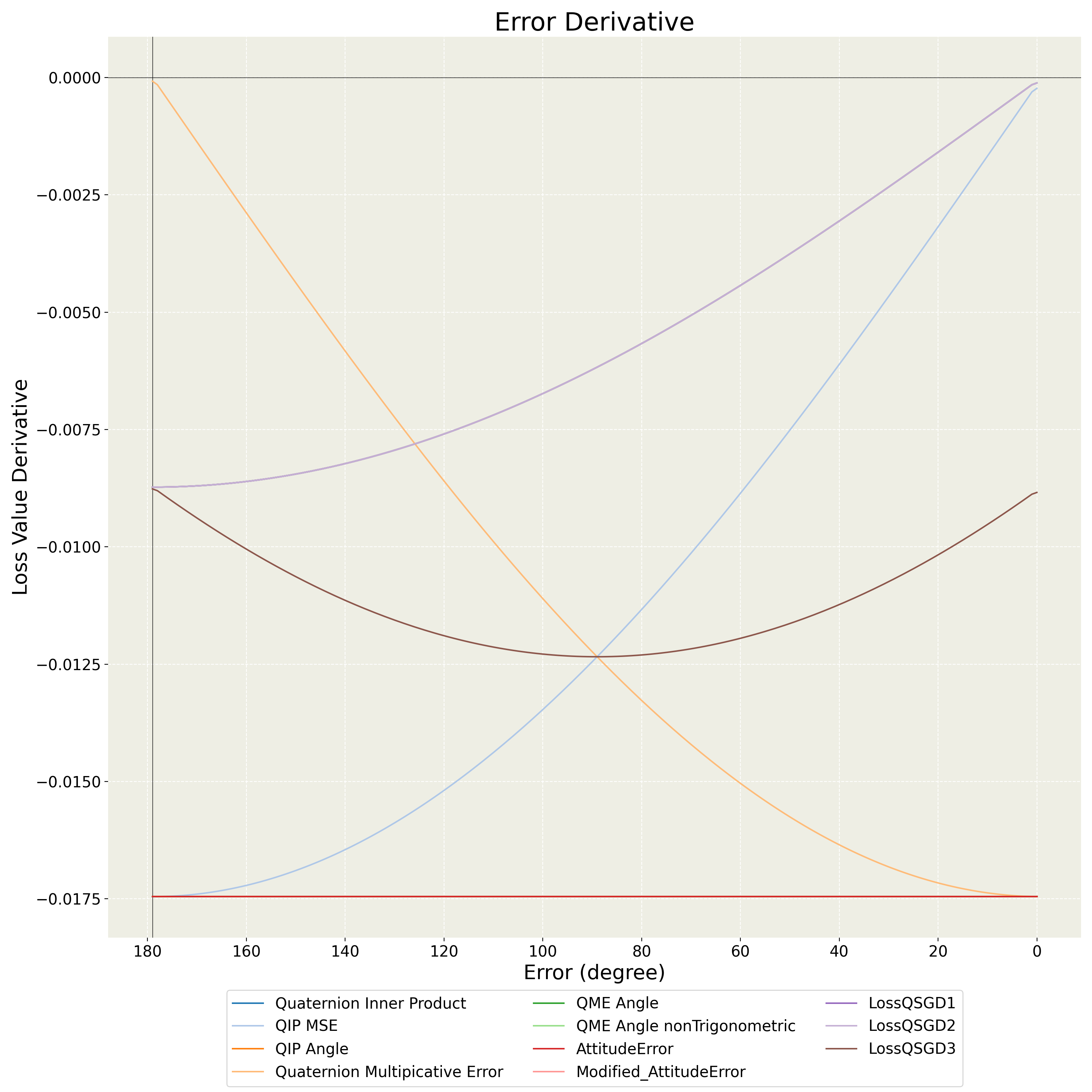}
    \caption{Compare Loss functions derivative for attitude error \cite{golroudbari2019design}}
\label{fig:loss_functionsDerv}
\end{figure}

Figures \ref{fig:loss_functions}, \ref{fig:loss_functionsDerv}, and \ref{fig:loss_functionsComp} show the loss values for the attitude error from $|\pi|$ to $0$.

In the context of attitude estimation, accurate determination of a vehicle's orientation is crucial in aerospace engineering. One widely used method for estimating orientation is through the use of quaternions, which represent rotation around a unit vector. Error matrices are essential in quantifying the accuracy of an estimation algorithm. These matrices relate the error in the estimated attitude to the error in the measured sensor data. Various types of error matrices can be used, depending on the specific application and desired properties of the estimated attitude. Therefore, it is important to carefully choose and fine-tune these parameters, including the training data and network architecture, to achieve optimal results. In the realm of deep learning, the loss function plays a critical role in the training process of a deep neural network by measuring the difference between predicted and actual output values. However, for geometric quantities such as attitude error, algebraic error matrices such as the MSE or MAE are unsuitable. Instead, several methods have been developed for defining and implementing quaternion error, such as the use of geodesic distances. The choice of the error representation method should be carefully considered to ensure accurate and reliable attitude estimation.

\subsection{Proposed Network Architecture}

Our research focuses on developing end-to-end deep learning frameworks for inertial attitude estimation, which can be generalized across various environments and sensor sampling rates. While there have been previous works that utilize deep learning for inertial odometry \cite{kim2021nine, silva2019end, esfahani2019aboldeepio}, visual-inertial odometry \cite{han2019deepvio, liu2021atvio, shamwell2019unsupervised}, visual attitude estimation \cite{phisannupawong2020vision}, and enhancing sensor fusion techniques accuracy \cite{al2019deep}, only three models have explored the use of end-to-end learning-based methods for inertial attitude estimation. The first model, RIANN \cite{weber2021riann}, introduced a GRU-based approach, but the lack of quantitative information in their article makes it difficult to reproduce their results. The second study \cite{narkhede2021incremental} presented an LSTM-based model, but it was limited to only one sampling rate, and the authors did not mention it in their publication or test it on publicly available inertial datasets. The third study \cite{brotchie2022leveraging} employed a self-attention mechanism, but their model was trained and tested only on the OxIOD dataset \cite{chen2018oxiod}, which limits its applicability to only one type of motion and a specific sampling rate.

Our contribution to the field is the introduction of a novel end-to-end learning framework for estimating orientation that is independent of external data sources and can generalize across various environments and sensor sampling rates. To achieve this, we utilized a combination of CNNs and LSTMs. Our model is trained and tested on publicly available datasets, demonstrating its ability to generalize across different sampling rates and environments. We evaluate the performance of our model against existing state-of-the-art methods and show that our approach achieves competitive accuracy with lower computational complexity. Our study aims to make a significant contribution to this critical area of research and pave the way for more accurate and robust attitude estimation in various applications, such as autonomous vehicles, drones, and robotics.
\par
The real-time attitude estimation problem can be addressed using several deep learning topologies. Studies have shown that RNNs, such as LSTM and GRU, CNN, and hybrid RNN-CNN networks can handle IMU data to estimate system state variables. The choice of topology must also consider the computational cost since real-time application is a primary requirement. Therefore, we evaluated GRU-based, LSTM-based, CNN-based, and hybrid CNN-based networks to estimate the system's attitude. Based on our results, we selected three different models, and their proposed network architectures are shown in fig.\ref{modelA}, fig.\ref{modelB}, and fig.~\ref{modelC}.

\begin{figure}[!ht]
\centering \includegraphics[width=\linewidth]{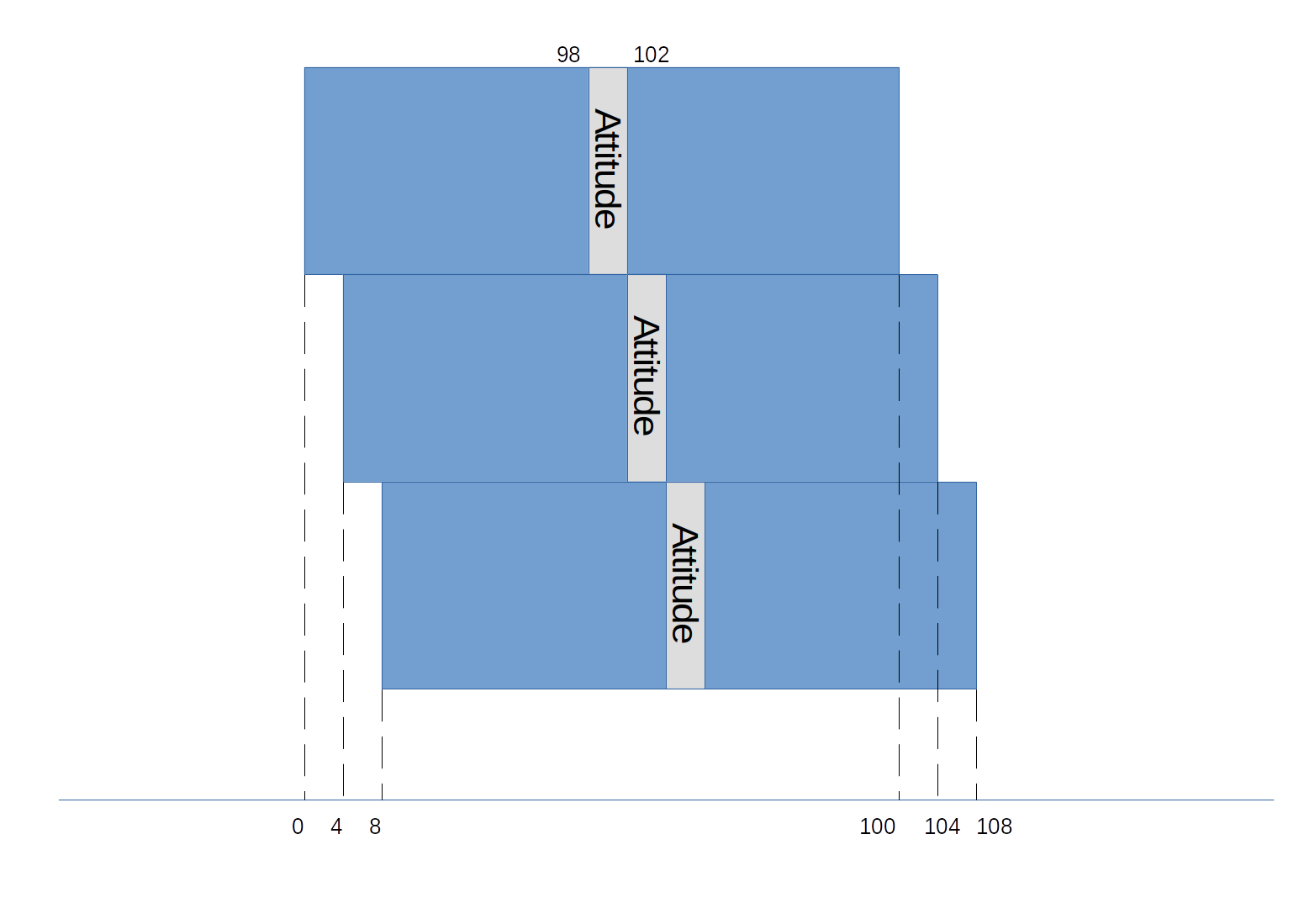}
\caption{Time window for attitude estimation, where both past and future data are used to estimate the attitude at each time step.}\label{time_window}
\end{figure}

Similar to \cite{chen2018ionet,silva2019end,kim2021nine,yan2018ridi,herath2020ronin}, we used windows of $\mathcal{N}$ frames as input data for the model, which contained 3-axis acceleration and 3-axis angular velocity. We used the past $\frac{\mathcal{N}}{2}$ and future $\frac{\mathcal{N}}{2}$ IMU measurements to predict the attitude. We also used a stride size of $\mathcal{S}$ between consecutive IMU measurements, and the attitude estimation occurred between $\frac{\mathcal{N}}{2} - \mathcal{S}$ and $\frac{\mathcal{N}}{2} + \mathcal{S}$ frames, as shown in fig.~\ref{time_window}.

Based on previous studies and existing models, we initially implemented our own model for inertial data handling. We drew inspiration from several sources, including \cite{weber2021riann,narkhede2021incremental,brotchie2022leveraging,chen2019deep,silva2019end}, and worked to modify and optimize our model for our specific application.

To fine-tune our model, we employed various hyperparameter optimization techniques such as random search, grid search, and population-based training algorithms. We tested different architectures, layers, and activation functions, as well as experimented with various batch sizes and learning rates. By analyzing the performance metrics of each model variant, we were able to select the best architecture for our specific use case.

Overall, our approach involved a combination of building upon previous research and utilizing advanced optimization techniques to create a model tailored to our specific needs. By iterating and improving upon our initial implementation, we were able to achieve high accuracy and performance in our inertial attitude estimation task.

The proposed network architecture consists of four main components: (1) Feature Extraction, (2) Feature Fusion, (3) Sampling Rate Fusion, and (4) Attitude Estimation. We added Gaussian noise layers to the model's inputs to improve generalization by adding random noise to the input data during training. This can make the model more robust to small variations in the input and prevent overfitting. It can also serve as a regularization technique to reduce overfitting by adding random noise to the inputs, which is sampled from a Gaussian distribution with standard variation of 0.25. We also used dropout layers in the feature extraction and feature fusion layers to improve the network's robustness.

The feature extraction component extracts the features from the IMU data. We split each axis of the IMU data in Model A and B and then fed each into a layer. After concatenating the layers, they are fed to a layer to fuse the extracted feature. To consider the sampling rate, the fused features are connected to the sampling rate layer. In the last layer, a feed-forward network with four units is followed by a unit scaling layer to estimate the attitude.

\subsubsection{Model A}
\begin{figure*}[!ht]
	\centering \includegraphics[width=\textwidth]{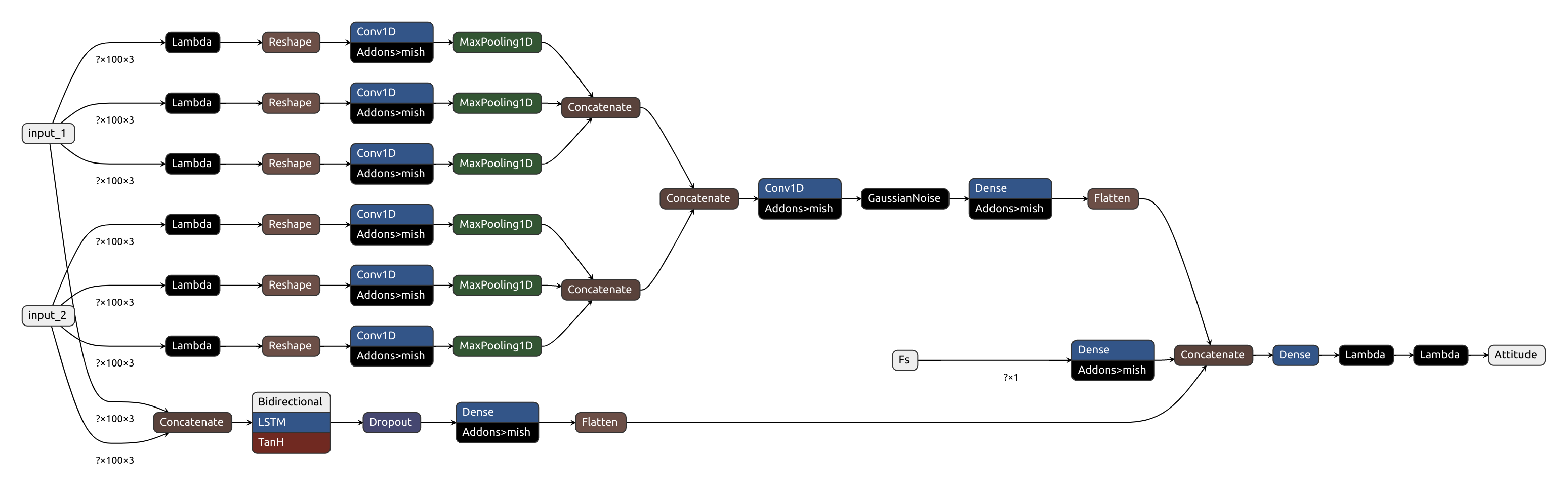}
	\caption{Proposed Network Architecture of Model A}\label{modelA}
\end{figure*}

The model that we developed for handling the inertial data consists of a one-dimensional CNN in combination with bidirectional long short-term memory, Model A, as shown in Figure~\ref{modelA}. The input data is fed into the model in windows of $\mathcal{N} = 100$ frames, containing 3-axis acceleration and 3-axis angular velocity. The past $50$ and future $50$ IMU measurements are used to predict the attitude. The consecutive IMU measurements have a stride size of $4$, and new attitude will be calculated in every $4$ frames. 

The feature extraction component extracts the features from the IMU data using separate 1D-CNN layers with a filter size of 128 and kernel size of 11 for each axis of the IMU data. Each CNN layer is followed by a max pooling layer with a pooling size of 3. The output of all the layers is concatenated and fed into a CNN layer with 128 filters, kernel size of 11, and stride size of 1. The CNN layer fuses the extracted features in the last layer and is followed by a feed-forward layer with 512 units. To take advantage of sequence modeling and temporal information, a bidirectional LSTM layer with 128 units is used. Dropout layers are used to reduce overfitting in the feature extraction and feature fusion layers.

The output of the LSTM layer is concatenated with the output of the CNN layer with Mish activation function, and the output of a fully connected layer with 512 neurons. The fully connected layer takes the sampling rate of the IMU measurements as input and works as the sampling rate fusion component, which is used to fuse the extracted features from the IMU data with different sampling rates. The Attitude Estimation component is composed of a fully connected layer with four neurons representing the system's estimated attitude in quaternion form. Gaussian noise layers are added to the model's inputs to help with generalization, by adding random noise to the input data during training. The noise is sampled from a Gaussian distribution with a standard variation of 0.25.

\subsubsection{Model B}
\begin{figure*}[!ht]
	\centering \includegraphics[width=\textwidth]{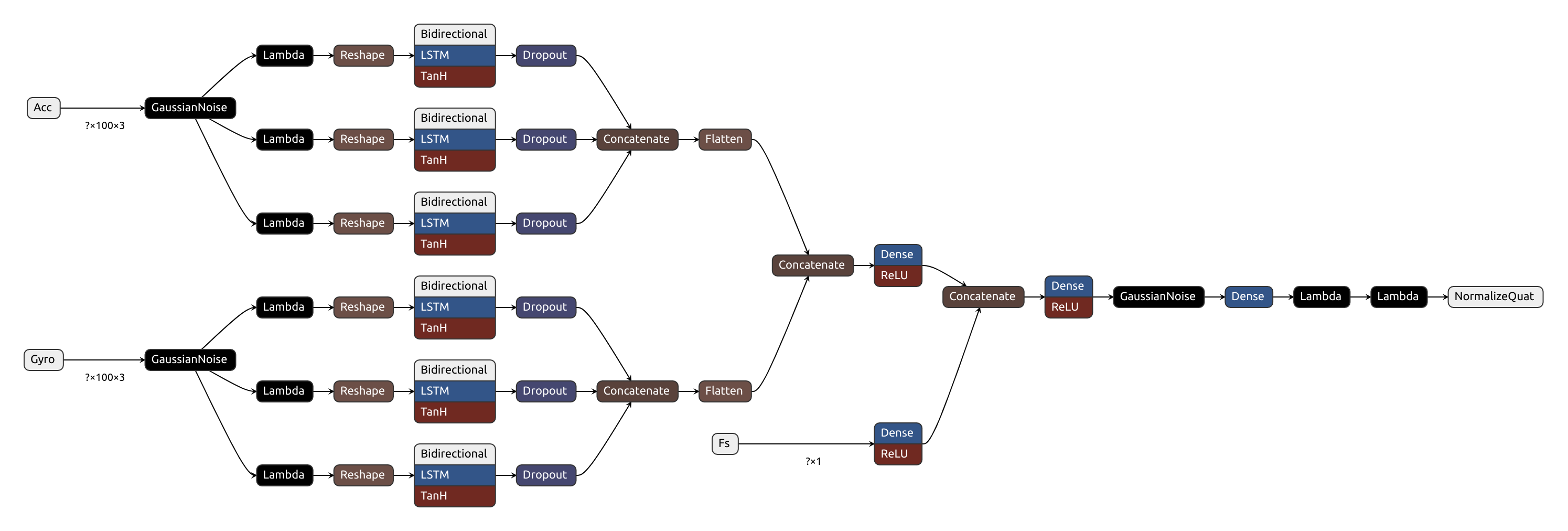}
	\caption{Proposed Network Architecture of Model B}\label{modelB}
\end{figure*}

Model B employs multiple Bidirectional LSTM layers with 50 units, each followed by a dropout layer. The Bidirectional LSTM layer's output is concatenated and passed into a feed-forward layer with 256 units and ReLU activation function. The sampling rate of IMU sensors is provided as input to a dense layer with 256 units and ReLU activation function. The outputs of the dense layers are concatenated and passed into a final dense layer with four units and linear activation function, followed by a unit scaling layer to estimate the quaternions. The proposed network architecture of Model B is depicted in Figure~\ref{modelB}.

\subsubsection{Model C}
\begin{figure*}[!ht]
	\centering \includegraphics[width=\textwidth]{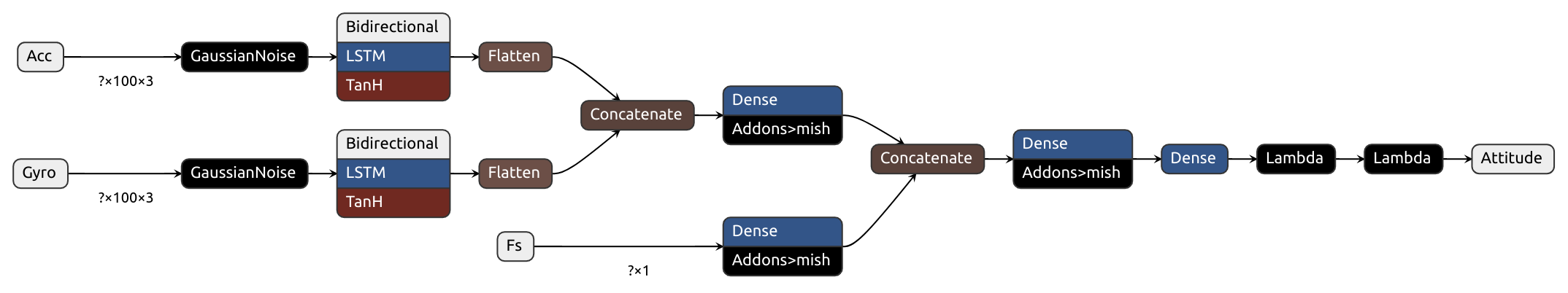}
	\caption{Proposed Network Architecture of Model C}\label{modelC}
\end{figure*}

In Model C, two Bi-LSTM layers are employed, followed by a dense layer with 256 units. The sampling rate is provided as input to a similar dense layer with Mish activation function. The output of the dense layer is concatenated and passed into another dense layer with 256 units. The output consists of a dense layer with four units and linear activation function, followed by a unit scaling layer to estimate the quaternions. The proposed network architecture of Model C is depicted in Figure~\ref{modelC}.

\subsection{Learning Rate Finder}

In this study, we utilized the learning rate finder technique to identify the optimal learning rate for training our machine learning model. The learning rate finder technique is a widely used method to determine the optimal learning rate for machine learning models, by gradually increasing the learning rate from a minimal value and monitoring the loss's behavior over time \cite{donini2019scheduling}. The optimal learning rate refers to the value at which the loss decreases at an appropriate rate without compromising the model's ability to explore different regions of the parameter space.

To use this technique, we executed multiple experiments with varying learning rates and monitored the performance on validation datasets. The best-performing experiment was then selected as the optimal learning rate for training the model. The learning rate finder technique provides an efficient and fast approach for model convergence compared to traditional methods, such as constant decay or exponential decay schedules.

Commonly used learning rate schedules include constant, exponential decay, step-wise decay, and cyclical learning rates (CLR). A constant schedule maintains the same learning rate for all training iterations, while an exponential decay schedule gradually reduces the learning rate over time. The idea of using an exponentially decreasing learning rate was first proposed by Leonid Khachiyan in 1980 \cite{khachiyan1980polynomial} to enhance the convergence speed and accuracy of gradient descent algorithms. The equation for computing the learning rate in this case is as follows:

\begin{equation}
\begin{gathered}
lr = lr * decay\_rate^{step/decay\_step},
\end{gathered}
\end{equation}

where $lr$ is the current learning rate, step is the current training iteration, and $decay\_step$ determines how frequently the learning rate is reduced.

Step-wise decays involve reducing the learning rate at specific intervals during the training process. In \cite{hinton2012improving}, G. Hinton proposed using a step-wise decay schedule to facilitate the learning rate at specific intervals during the training process. This technique has become a popular method for improving model performance and avoiding local minima. The learning rate in this case can be computed using the following equation:

\begin{equation}
\begin{gathered}
lr = lr * factor \\
\text{or} \\
lr = lr - fixed\ amount.
\end{gathered}
\end{equation}

CLR involves gradually increasing and decreasing the learning rate over time, enabling the model to explore different regions of the parameter space more efficiently \cite{smith2017cyclical}. This is done by setting upper and lower bounds for the range of values that can be explored and a step size that determines how quickly or slowly the value changes between these bounds. The cyclical learning rate approach allows the model to avoid local minima while still converging on an optimal solution faster than traditional methods such as constant or exponential decay schedules. The equation for computing the cyclical learning rate is:

\begin{equation}
\begin{gathered}
lr = \frac{lower_bound}{2} + \frac{upper_bound}{2} * (1 + cos(step/stepsize)),
\end{gathered}
\end{equation}

where $lr$ is the current learning rate, step is the current training iteration, and step size determines how quickly or slowly the value changes between the upper and lower bounds.

\begin{figure}[!ht]
	\centering \includegraphics[width=0.5\linewidth]{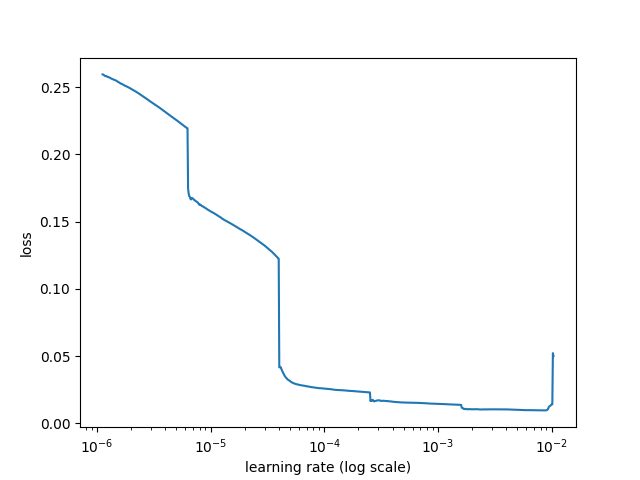}
	\caption{Learning Rate Finder}\label{fig2}
\end{figure}

In this study, we utilized the CLR method to determine the optimal learning rate. We trained the network for several epochs and plotted the loss value against the learning rate. We chose the learning rate value that had the steepest gradient in the loss curve (Fig.~\ref{fig2}).

\section{Experiment}
\subsection{Dataset}
\subsubsection{Introduction}
In the field of attitude estimation algorithms, IMU datasets play a vital role in evaluating algorithm performance. To evaluate and compare various attitude estimation algorithms, we require datasets that consist of IMU measurements. IMU datasets can be categorized into two categories: synthetic and real-world. Synthetic datasets are generated by simulating IMU measurements. In contrast, real-world datasets are collected from actual experiments. Real-world experiments can further be categorized as indoor and outdoor experiments, depending on whether they are conducted in a controlled environment, such as a laboratory, or an uncontrolled environment, such as a car.

To train, validate and test any neural network model, a comprehensive and accurate database is required. The performance of the Deep Learning model is directly affected by the quality of data used for its training. Therefore, we require a dataset containing both input and output parameters with certain conditions:
\begin{itemize}
\item The input and output parameters must be accurate and reliable.
\item The dataset must be of sufficient size to train the Deep Learning model effectively.
\item The dataset must be diverse enough to cover all possible scenarios.
\end{itemize}

Next, we will present some of the most commonly used IMU datasets, which satisfy the above conditions.

\subsubsection{RepoIMU T-stic}
The RepoIMU T-stick \cite{szczkesna2016reference} is an affordable, small, and high-performance IMU designed for versatile applications. It is a 9-axis sensor that measures acceleration, angular velocity, and magnetic field. The dataset comprises two sets of experiments recorded with a T-stick and a pendulum. The T-stick data includes 29 trials, each lasting approximately 90 seconds. The IMU is attached to a T-shaped stick with six reflective markers, and each experiment involves slow or fast rotation around a principal sensor axis or translation along a principal sensor axis. The Vicon Nexus OMC system and XSens MTi IMU data are synchronized and provided at a frequency of 100 Hz. The authors have disclosed that the IMU coordinate system and the ground trace are not aligned and have proposed a quaternion-based approach to compensate for one of the two required rotations. However, some experiments contain gyroscope clipping and ground tracking issues, significantly impacting the resulting errors. Therefore, preprocessing and exclusion of some trials may be necessary while evaluating the model's accuracy using this dataset.

\subsubsection{RepoIMU T-pendulum}
The second part of the RepoIMU dataset \cite{szczkesna2016reference} contains data from a triple pendulum with IMUs mounted on it. The measurement data is provided at 90 Hz or 166 Hz, but the IMU data has duplicate samples, likely due to artificial sampling or transmission problems where missed samples are replaced by duplicating the last sample received, reducing the effective sampling rate. When discarding frequent samples, the sampling rate achieved is about 25 Hz and 48 Hz for the accelerometer and gyroscope, respectively. Due to this issue, using this dataset for model training and evaluation is not recommended, and it is not suitable for evaluating the accuracy of Inertial Orientation Estimation (IOE) with high precision.

\subsubsection{Sassari}
The Sassari dataset \cite{caruso2020orientation} aims to validate a parameter tuning approach based on the orientation difference of two IMUs of the same model. The dataset includes six IMUs from three manufacturers (Xsens, APDM, Shimmer) placed on a wooden board. Rotation around specific axes and free rotation around all axes are repeated at three different speeds. Data is synchronized and presented at 100 Hz, and local coordinate frames are aligned by precise manual placement. The dataset includes 18 experiments (3 speeds, 3 IMU models, and 2 IMUs of each model). Although the dataset contains different speeds and types of IMUs, it lacks robust variation in the type of movement and magnetic data diversity, as all motions occur in a homogeneous magnetic field and do not involve pure translational motions. Therefore, the model trained with this dataset may not be general and robust. However, it can be used for evaluating the model. The total movement duration of all three trials is 168 seconds, with the longest movement phase lasting 30 seconds, making it unsuitable for training due to its short duration.

\subsubsection{The Oxford Inertial Odometry Dataset}
The Oxford Inertial Odometry Dataset (OxIOD) \cite{chen2018oxiod} is a comprehensive and expansive collection of inertial data captured by smartphones, mainly using the iPhone 7 Plus model, at a high sampling rate of 100 Hz. The dataset comprises 158 tests that span a distance of over 42 km, with OMC ground tracks available for 132 of these tests. The primary objective of this dataset is to facilitate the evaluation of inertial odometry models. As such, it excludes pure rotational and translational movements that could provide a systematic evaluation of the model's performance under different scenarios. Nonetheless, it covers a diverse range of everyday activities.

Despite its considerable utility, OxIOD lacks comprehensive descriptions of certain essential parameters such as the alignment of the coordinate frames. Moreover, the orientation of the ground trace exhibits frequent irregularities, including abrupt changes in orientation unaccompanied by similar changes in the IMU data. To maximize the utility of the dataset, it is necessary to recognize these limitations and account for them appropriately in any analysis or model evaluation.

The OxIOD dataset can serve as a valuable resource for researchers working in inertial odometry, mobile robotics, and related fields. With its rich collection of inertial data and comprehensive test suite, it provides a platform for developing and evaluating advanced inertial odometry models. 

\subsubsection{MAV Dataset}
Most datasets suitable for the simultaneous localization and mapping problem are collected from sensors such as wheel encoders and laser range finders mounted on ground robots. For small air vehicles, there are few datasets, and MAV Dataset \cite{lee2010benchmarking} is one of them. This data set was collected from the sensor array installed on the "Pelican" quadrotor platform in an environment. The sensor suite includes a forward-facing camera, a downward-facing camera, an inertial measurement unit, and a Vicon ground-tracking system. Five synchronized datasets are presented
\begin{itemize}
	\item 1LoopDown
	\item 2LoopsDown
	\item 3LoopsDown
	\item hoveringDown
	\item randomFront
\end{itemize}
These datasets include camera images, accelerations, heading rates, absolute angles from the IMU, and ground tracking from the Vicon system.
\subsubsection{EuRoC MAV}
The EuRoC MAV dataset \cite{burri2016euroc} is a large dataset collected from a quadrotor MAV. The dataset contains the internal flight data of a small air vehicle (MAV) and is designed to reconstruct the visual-inertial 3D environment. The six experiments performed in the chamber and synchronized and aligned using the OMC-based Vicon ground probe are suitable for training and evaluating the model's accuracy. It should be noted that camera images and point clouds are also included.
This set does not include magnetometer data, which limits the evaluation of three degrees of freedom and is only for two-way models (including accelerometer and gyroscope). Due to the nature of the data, most of the movement consists of horizontal transfer and rotation around the vertical axis. This slope does not change much during the experiments. For this reason, it does not have a suitable variety for model training. Since flight-induced vibrations are visible in the raw accelerometer data, the EuRoC MAV dataset provides a unique test case for orientation estimation with perturbed accelerometer data.
\subsubsection{TUM-VI}
The TUM Visual-Inertial dataset \cite{schubert2018tum} is suitable for optical-inertial odometry and consists of 28 experiments with a handheld instrument equipped with a camera and IMU. Due to this application focus, most experiments only include OMC ground trace data at the experiment's beginning and end. However, the six-chamber experiments have complete OMC data. They are suitable for evaluating the accuracy of the neural network model. Similar to the EuRoC MAV data, the motion consists mainly of horizontal translation and rotation about the vertical axis, and magnetometer data is not included.
\subsubsection{KITTI}
The KITTI Vision Benchmark Suite \cite{geiger2012we} is a large set of data collected from a stereo camera and a laser range finder mounted on a car. The dataset includes 11 sequences with a total of 20,000 images. The dataset is suitable for evaluating the model's accuracy in the presence of optical flow. However, the dataset does not include magnetometer data, which limits the evaluation of three degrees of freedom and is only for two-way models (including accelerometer and gyroscope).
\subsubsection{RIDI}
RIDI datasets \cite{yan2018ridi} were collected over 2.5 hours on ten human subjects using smartphones equipped with a 3D tracking capability to collect IMU-motion data placed on four different surfaces (e.g., the hand, the bag, the leg pocket, and the body). The Visual Inertial SLAM technique produced the ground-truth motion data. They recorded linear accelerations, angular velocities, gravity directions, device orientations (via Android APIs), and 3D camera poses with a Google Tango phone, Lenovo Phab2 Pro. Visual Inertial Odometry on Tango provides camera poses that are accurate enough for inertial odometry purposes (less than 1 meter after 200 meters of tracking).
\subsubsection{RoNIN}
The RoNIN dataset \cite{herath2020ronin} contains over 40 hours of IMU sensor data from 100 human subjects with 3D ground-truth trajectories under natural human movements. This data set provides measurements of the accelerometer, gyroscope, dipstick, GPS, and ground track, including direction and location in 327 sequences and at a frequency of 200 Hz. A two-device data collection protocol was developed. A harness was used to attach one phone to the body for 3D tracking, allowing subjects to control the other phone to collect IMU data freely. It should be noted that the ground track can only be obtained using the 3D tracker phone attached to the harness. In addition, the body trajectory is estimated instead of the IMU.
\subsubsection{BROAD}
The Berlin Robust Orientation Evaluation (BROAD) dataset \cite{laidig2021broad} includes a diverse set of experiments covering a variety of motion types, velocities, undisturbed motions, and motions with intentional accelerometer perturbations as well as motions performed in the presence of magnetic perturbations. This data set includes 39 experiments (23 undisturbed experiments with different movement types and speeds and 16 experiments with various intentional disturbances). The data of the accelerometer, gyroscope, magnetometer, quaternion, and ground tracks, are provided in an ENU frame with a frequency of 286.3 Hz.

\subsection{Training}

To build our attitude estimation models, we utilized the Lima, Kim, and Chen models, which have been established as a reliable infrastructure for our purposes \cite{silva2019end,kim2021nine,chen2018oxiod}. Our proposed method employs deep learning techniques to estimate attitude based on inertial measurements. The models take a sequence of accelerometer and gyroscope readings along with their respective timestamps as input and produces roll and pitch angles as output. By using an end-to-end deep learning framework, the models is capable of handling noise and bias inherent in IMU measurements.

Our models incorporates a combination of CNN and LSTM layers. The CNN layers are responsible for feature extraction from the accelerometer and gyroscope readings, while the LSTM layers are used to learn temporal dependencies between the extracted features. The input to the network consists of a sequence of 100 accelerometer and gyroscope readings. 

To prevent over-fitting, we added a dropout layer with a 0.25 probability after each LSTM layer. This layer randomly drops out $25\%$ of the units in the layer during training. The input in each time step is a window of 100 accelerometer and gyroscope readings which consists of 50 past and 50 future readings. The window's stride is two frames, leading the model to estimate the attitude every two frames.

We trained the network using the Adam optimizer with a initial learning rate of 0.00156 and the Quaternion Multiplicative Error loss function. The training was performed on the combination datasets (Tab ~\ref{Train}) for 500 epochs with a batch size of 500. We implemented the network using the Keras library with the TensorFlow backend.

\begin{table}[!ht]
  \caption{Training datasets used in the study.\label{Train}}
  \centering
  \begin{tabular}{ll}
  \hline
  \textbf{Dataset} & \textbf{Sequence Numbers} \\
  \hline
  BROAD & 1-8, 12, 15-18, 20-23, 26, 28-30, 38, 39 \\
  OxIOD & 1-12 \\
  RepoIMU TStick & 1-13 \\
  Sassari & 1-9 \\
  RIDI & 1-20 \\
  RONIN & 1-12 \\
  \hline
  \end{tabular}
  \end{table}

In the training phase of the deep learning model, it is essential to carefully select the datasets to ensure that the model can capture the complexity of various scenarios and conditions. The choice of datasets can impact the performance of the model, and therefore, it is necessary to choose datasets that represent a range of motion patterns and environmental perturbations.

In this study, we selected several datasets to use in the training phase, namely BROAD, OxIOD, RepoIMU TStick, Sassari, RONIN, and RIDI. These datasets provide various types of motion patterns, such as rotation and translation, and are recorded under different environmental conditions, such as indoor and outdoor settings, to ensure that the trained model is robust and can generalize well to unseen scenarios.

Overall, the training dataset used in this study consisted of 89 trials, of which 20\% was used for validation, and 399 trials were selected for testing. By using a diverse range of datasets and a sufficient number of trials, we aimed to train a deep learning model that can accurately estimate the attitude of IMU sensors under various scenarios and sampling rates.

\subsection{Evaluation}
In this section, we present the extensive evaluation of the proposed end-to-end deep-learning approaches for real-time attitude estimation using inertial sensor measurements. Our evaluation involved six publicly available datasets, which provide a wide range of motion patterns, sampling rates, and environmental disturbances. To the best of our knowledge, this is the most comprehensive benchmark conducted on this problem.

We conducted multiple runs of the experiments and averaged the results to ensure that the evaluation was representative of the performance of the proposed method. This helped reduce the influence of random fluctuations or noise in the data and provided a more accurate representation of the method's performance.

Our evaluation demonstrated that the proposed method is effective and reliable for real-time attitude estimation using inertial sensor measurements. The evaluation results showed that the proposed method outperformed other state-of-the-art approaches in terms of accuracy and robustness and exhibited strong generalization capabilities over a wide range of motion patterns, sampling rates, and environmental disturbances.

Specifically, our proposed method outperformed the state-of-the-art methods, including Madgwick, Mahony, CF, and RIANN, in terms of accuracy and robustness. The RMSE and QE values were consistently lower for the proposed method, indicating a higher level of accuracy in the attitude estimates.

In Tables~\ref{table_sassari}, \ref{table_broad}, \ref{table_oxiod}, \ref{table_ronin}, \ref{table_ridi}, and \ref{table_repo} below, we present the evaluation results of the proposed method and the other approaches on each dataset. Our evaluation results demonstrate that the proposed method is effective and reliable for real-time attitude estimation using inertial sensor measurements and outperforms the state-of-the-art methods in terms of accuracy and robustness.

\begin{table*}[!ht]
\caption{Evaluation results of the proposed method and the other approaches on the RIDI dataset. \label{table_ridi}}
\centering
\begin{tabular}{lccccccc}
\hline
\textbf{Trial No,} & \textbf{Model A} & \textbf{Model B} & \textbf{Model C} &  \textbf{RIANN} & \textbf{CF} & \textbf{Madgwick} & \textbf{Mahony} \\
\hline
\textbf{Av. Dan} & 0.84 & 0.58 & 0.72 & 1.20 & 8.79& 1.94 & 2.29 \\
\textbf{Av. Hang} & 1.40 & 1.34 & 1.39 & 1.28 & 7.33& 1.99 & 2.00 \\
\textbf{Av. Hao} & 3.05 & 2.81 & 2.85 & 1.53 & 9.27& 1.91 & 2.44 \\
\textbf{Av. Huayi} & 2.74 & 2.58 & 2.60 & 1.30 & 8.90& 2.01 & 2.35 \\
\textbf{Av. Ma} & 2.87 & 2.61 & 2.49 & 1.32 & 5.19& 2.03 & 1.74 \\
\textbf{Av. Ruixuan} & 2.58 & 2.68 & 2.63 & 1.54 & 8.32 & 2.17 & 2.34 \\
\textbf{Av. Shali} & 2.34 & 2.17 & 2.21 & 1.15 & 8.56 & 1.85 & 2.32 \\
\textbf{Av. Tang} & 3.14 & 2.87 & 3.05 & 1.50 & 8.59 & 2.43 & 2.17 \\
\textbf{Av. Xiaojing} & 2.21 & 2.16 & 2.16 & 1.23 & 6.40 & 2.28& 1.83 \\
\textbf{Av. Yajie} & 2.35 & 2.31 & 2.33 & 1.46 & 7.05 & 2.10 & 1.96 \\
\textbf{Av. Zhicheng} & 2.54 & 2.29 & 2.20 & 1.32 & 8.15 & 2.31 & 2.08 \\
\hline
\textbf{Average All} & 2.17 & 2.03 & 2.06 & 1.34 & 7.85 & 2.07 & 2.13\\
\hline
\end{tabular}
\end{table*}

\begin{table*}[!ht]
\caption{Evaluation results of the proposed method and the other approaches on the RepoIMU TStick dataset. \label{table_repo}}
\centering
\begin{tabular}{lccccccc}
\hline
\textbf{Trial No,} & \textbf{Model A} &  \textbf{Model B} & \textbf{Model C} &  \textbf{RIANN} & \textbf{CF} & \textbf{Madgwick} & \textbf{Mahony} \\
\hline
\textbf{Av. Test 2}& 0.84 & 0.49 & 0.71 & 2.25 & 3.58 & 1.65& 1.73\\
\textbf{Av. Test 3}& 1.05 & 0.73 & 1.08 & 4.96 & 5.32 & 4.38& 4.38\\
\textbf{Av. Test 4}& 1.09 & 0.69 & 0.85 & 2.28 & 2.26 & 2.28& 2.30\\
\textbf{Av. Test 5}& 9.03 & 3.96 & 7.39 & 52.78 & 26.63& 72.97 & 40.74 \\
\textbf{Av. Test 6}& 3.00 & 1.32 & 1.69 & 4.95 & 28.75& 6.00& 10.72 \\
\textbf{Av. Test 7}& 4.76 & 4.29 & 3.64 & 3.31 & 16.90& 2.90& 4.72\\
\textbf{Av. Test 8}& 6.86 & 3.93 & 3.30 & 1.69 & 9.16 & 1.86& 3.49\\
\textbf{Av. Test 9}& 5.08 & 4.15 & 2.88 & 2.08 & 11.51& 2.15& 2.93\\
\textbf{Av. Test 10} & 9.13 & 5.05 & 4.21 & 3.16 & 8.64 & 4.39& 2.97\\
\textbf{Av. Test 11} & 5.97 & 5.42 & 5.27 & 3.40 & 6.31 & 3.60& 3.73\\
\hline
\textbf{Av. All} & 5.07 & 3.28 & 3.36 & 8.72 & 11.98& 11.09& 8.19 \\
\hline
\end{tabular}
\end{table*}

\begin{table*}[!ht]
\caption{Evaluation results of the proposed method and the other approaches on the Sassari dataset.\label{table_sassari}}
\centering
\begin{tabular}{lccccccc}
\hline
\textbf{Trial No,} & \textbf{Model A} & \textbf{Model B} & \textbf{Model C} &  \textbf{RIANN} & \textbf{CF} & \textbf{Madgwick} & \textbf{Mahony} \\
\hline
\textbf{fast\_v4\/AP1} & 0.78 & 0.57 & 0.80 & 1.82& 5.36 & 1.76 & 2.19 \\
\textbf{fast\_v4\/AP2} & 0.88 & 0.70 & 0.65 & 1.38& 5.35 & 1.47 & 2.00 \\
\textbf{fast\_v4\/SH1} & 2.37 & 1.75 & 1.79 & 4.16& 7.76 & 4.40 & 3.94 \\
\textbf{fast\_v4\/SH2} & 7.48 & 4.12 & 6.26 & 14.49 & 14.39& 14.29& 14.37\\
\textbf{fast\_v4\/XS1} & 0.67 & 0.46 & 0.56 & 2.34 & 4.46 & 2.13 & 2.07 \\
\textbf{fast\_v4\/XS2} & 0.81 & 0.65 & 0.65 & 1.19 & 4.78 & 1.23 & 1.73 \\
\textbf{medium\_v4\/AP1} & 1.01 & 0.64 & 0.88 & 1.35 & 3.74 & 1.33 & 1.78 \\
\textbf{medium\_v4\/AP2} & 0.73 & 0.46 & 0.61 & 1.47 & 3.36 & 1.29 & 1.62 \\
\textbf{medium\_v4\/SH1} & 2.17 & 1.50 & 1.78 & 5.02 & 6.82 & 5.00 & 4.54 \\
\textbf{medium\_v4\/SH2} & 18.27 & 18.22 & 18.28 & 18.71 & 18.78& 18.62& 18.51\\
\textbf{medium\_v4\/XS1} & 1.64 & 1.41 & 1.40 & 1.83& 3.01 & 1.53 & 1.57 \\
\textbf{medium\_v4\/XS2} & 1.56 & 1.40 & 1.47 & 1.04& 2.98 & 1.10 & 1.34 \\
\textbf{slow\_v4\/AP1} & 2.29 & 2.60 & 2.25 & 1.23& 1.80 & 0.90 & 1.28 \\
\textbf{slow\_v4\/AP2} & 2.07 & 2.48 & 2.52 & 1.30 & 1.65 & 0.77 & 1.19 \\
\textbf{slow\_v4\/SH1} & 3.90 & 4.30 & 4.12 & 3.78 & 3.72 & 3.72 & 3.63 \\
\textbf{slow\_v4\/SH2} & 18.47 & 18.91 & 18.79 & 18.36 & 18.40& 18.31& 18.28\\
\textbf{slow\_v4\/XS1} & 2.01 & 2.61 & 2.42 & 2.10& 1.51 & 0.90 & 1.41 \\
\textbf{slow\_v4\/XS2} & 2.29 & 2.67 & 2.60 & 1.00& 1.64 & 0.81 & 1.04 \\
\hline
\textbf{ Average} &  3.86 & 3.64 & 3.77 & 4.59 & 6.08 & 4.42 & 4.58 \\
\hline
\end{tabular}
\end{table*}

\begin{table*}
\caption{Evaluation results of the proposed method and the other approaches on the BROAD dataset.\label{table_broad}}
\centering
\begin{tabular}{lccccccc}
\hline
\textbf{Trial No,} & \textbf{Model A} & \textbf{Model B} & \textbf{Model C} &  \textbf{RIANN} & \textbf{CF} & \textbf{Madgwick} & \textbf{Mahony} \\
\hline
\textbf{Trial No, 1}& 0.79 & 0.36 & 0.53 & 1.40& 5.62 & 1.29 & 0.85 \\
\textbf{Trial No, 2}& 0.81 & 0.37 & 0.55 & 0.52& 3.61 & 0.46 & 0.41 \\
\textbf{Trial No, 3}& 0.80 & 0.36 & 0.54 & 0.75& 5.25 & 0.69 & 0.67 \\
\textbf{Trial No, 4}& 0.64 & 0.28 & 0.42 & 1.84& 3.06 & 2.61& 0.89\\
\textbf{Trial No, 5}& 0.65 & 0.34 & 0.48 & 0.40 & 1.74& 0.35& 0.30\\
\textbf{Trial No, 6}& 0.87 & 0.36 & 0.53 & 0.98 & 6.54& 1.83& 1.16\\
\textbf{Trial No, 7}& 0.94 & 0.50 & 0.67 & 0.91 & 8.52& 1.22& 1.09\\
\textbf{Trial No, 8}& 0.89 & 0.42 & 0.55 & 2.71 & 14.07 & 12.60 & 2.62\\
\textbf{Trial No, 9}& 2.77 & 2.47 & 2.35 & 0.73 & 5.29& 0.68& 0.72\\
\textbf{Trial No, 10} & 3.71 & 3.56 & 3.64 & 0.35 & 6.42& 0.76& 2.20\\
\textbf{Trial No, 11} & 3.32 & 3.14 & 3.24 & 0.48 & 4.86& 1.01& 1.88\\
\textbf{Trial No, 12} & 1.85 & 1.04 & 1.58 & 0.59 & 3.18& 0.81& 1.40\\
\textbf{Trial No, 13} & 1.08 & 1.07 & 1.03 & 0.48 & 1.58& 0.71& 0.77\\
\textbf{Trial No, 14} & 1.62 & 1.62 & 1.59 & 0.40 & 2.32& 0.59& 0.90\\
\textbf{Trial No, 15} & 1.08 & 0.39 & 0.62 & 0.80 & 26.61 & 3.68& 5.09\\
\textbf{Trial No, 16} & 1.11 & 0.49 & 0.64 & 0.70 & 30.04 & 2.60& 7.43\\
\textbf{Trial No, 17} & 0.97 & 0.37 & 0.54 & 1.14 & 25.96 & 2.44& 5.26\\
\textbf{Trial No, 18} & 0.83 & 0.42 & 0.55 & 0.78 & 26.91 & 1.71& 10.26 \\
\textbf{Trial No, 19} & 2.15 & 2.08 & 1.90 & 1.43 & 3.57& 1.92& 1.63\\
\textbf{Trial No, 20} & 1.13 & 0.47 & 0.74 & 0.57 & 4.04& 0.95& 1.46\\
\textbf{Trial No, 21} & 1.22 & 0.52 & 0.77 & 3.23 & 32.65 & 20.20 & 8.29\\
\textbf{Trial No, 22} & 1.35 & 0.53 & 0.90 & 1.50 & 24.03 & 5.24& 5.42\\
\textbf{Trial No, 23} & 1.54 & 0.57 & 0.96 & 1.45 & 26.20 & 5.91& 6.94\\
\textbf{Trial No, 24} & 3.52 & 3.28 & 2.87 & 0.98 & 6.93& 1.15& 0.91\\
\textbf{Trial No, 25} & 3.61 & 3.44 & 3.49 & 0.62 & 5.91& 1.16& 1.92\\
\textbf{Trial No, 26} & 0.99 & 0.63 & 0.73 & 0.68 & 18.28 & 3.01& 1.33\\
\textbf{Trial No, 27} & 3.58 & 3.52 & 3.50 & 0.62 & 4.60& 2.19& 1.88\\
\textbf{Trial No, 28} & 1.34 & 0.58 & 0.81 & 2.96 & 24.18 & 12.12 & 5.26\\
\textbf{Trial No, 29} & 1.34 & 0.55 & 0.85 & 3.54 & 28.64 & 16.21 & 6.92\\
\textbf{Trial No, 30} & 1.03 & 0.51 & 0.74 & 1.63 & 28.62 & 9.84& 7.08\\
\textbf{Trial No, 31} & 3.83 & 3.46 & 3.51 & 1.54 & 22.56 & 9.61& 4.79\\
\textbf{Trial No, 32} & 2.41 & 2.36 & 2.34 & 0.44 & 5.47& 0.74& 2.11\\
\textbf{Trial No, 33} & 2.32 & 2.23 & 2.21 & 0.38 & 5.57& 0.80& 2.12\\
\textbf{Trial No, 34} & 2.32 & 2.27 & 2.19 & 0.59 & 6.14& 1.05& 2.26\\
\textbf{Trial No, 35} & 2.10 & 2.15 & 2.07 & 1.63 & 6.05& 5.36& 2.51\\
\textbf{Trial No, 36} & 2.68 & 2.91 & 2.69 & 0.68 & 8.91& 1.42& 2.59\\
\textbf{Trial No, 37} & 3.58 & 3.79 & 3.57 & 1.34 & 8.69& 5.82& 2.79\\
\textbf{Trial No, 38} & 1.69 & 0.58 & 1.02 & 0.75 & 9.25& 1.47& 2.89\\
\textbf{Trial No, 39} & 0.89 & 0.43 & 0.64 & 0.92 & 10.58 & 1.20& 2.76\\
\hline
\textbf{Average}& 3.86 & 3.64 & 3.77 & 4.59 & 6.08& 4.42& 4.58 \\
\hline
\end{tabular}
\end{table*}

\begin{table*}[!ht]
\caption{Evaluation results of the proposed method and the other approaches on the RoNIN dataset. \label{table_ronin}}
\centering
\begin{tabular}{lccccccc}
\hline
\textbf{Trial No,}& \textbf{Model A} & \textbf{Model B} & \textbf{Model C} & \textbf{RIANN} & \textbf{CF} & \textbf{Madgwick} & \textbf{Mahony} \\
\hline
\textbf{Av. train\_dataset\_1} & 6.07 & 5.59 & 5.63 & 1.75 & 13.95 & 2.55& 3.56\\
\textbf{Av. train\_dataset\_2} & 5.71 & 5.24 & 5.36 & 1.61 & 12.06 & 2.23& 3.22\\
\textbf{Av. seen\_subjects\_test\_set} & 4.82 & 4.49 & 4.72 & 1.80 & 14.87 & 2.72& 3.92\\
\textbf{Av. unseen\_subjects\_test\_set} & 6.02 & 5.70 & 5.80 & 1.67 & 13.65 & 2.26& 3.65\\
\hline
\textbf{Average All}& 5.69 & 5.28 & 5.39 & 1.71 & 13.66 & 2.46& 3.58 \\
\hline
\end{tabular}
\end{table*}

\begin{table*}[!ht]
\caption{Evaluation results of the proposed method and the other approaches on the OxIOD dataset. \label{table_oxiod}}
\centering
\begin{tabular}{lccccccc}
  \hline
\textbf{Trial No,}& \textbf{Model A} & \textbf{Model B} & \textbf{Model C} & \textbf{RIANN} & \textbf{CF} & \textbf{Madgwick} & \textbf{Mahony} \\
\hline
\textbf{Av. handbag}& 1.41 & 1.17 & 1.30 & 13.04 & 9.30& 12.88& 11.49\\
\textbf{Av. handheld }& 1.96 & 1.94 & 1.96 & 6.74& 3.87& 5.90 & 5.19 \\
\textbf{Av. iPhone 5}& 4.13 & 4.47 & 4.56 & 11.10 & 7.20& 11.20& 9.53 \\
\textbf{Av. iPhone 6}& 3.89 & 4.15 & 4.81 & 10.35 & 6.59& 10.30& 8.89 \\
\textbf{Av. user2}& 3.77 & 4.36 & 4.63 & 11.82 & 6.78& 11.79& 9.82 \\
\textbf{Av. user3 } & 4.87 & 5.58 & 6.21 & 12.62 & 8.23& 13.36& 10.82\\
\textbf{Av. user4 } & 5.20 & 5.40 & 5.51 & 13.08 & 7.56& 13.71& 10.98\\
\textbf{Av. user5}& 4.55 & 6.10 & 6.27 & 12.77 & 8.47& 12.91& 11.31\\
\textbf{Av. pocket} & 7.85 & 9.12 & 10.32& 15.10 & 10.01 & 16.02& 14.46\\
\textbf{Av. running}& 4.48 & 4.14 & 3.81 & 10.93 & 6.35& 10.58& 8.00 \\
\textbf{Av. slow walking} & 2.04 & 2.43 & 1.98 & 4.75& 3.43& 4.15 & 4.14 \\
\textbf{Av. trolley}& 4.52 & 5.31 & 4.66 & 4.71& 4.69& 4.69 & 4.70 \\
\hline
\textbf{Average All } & 3.92 & 4.37 & 4.51 & 10.01 & 6.49& 9.96 & 8.60 \\
\hline
\end{tabular}
\end{table*}

\begin{figure}[!ht]
\centering
\includegraphics[width=0.45\textwidth]{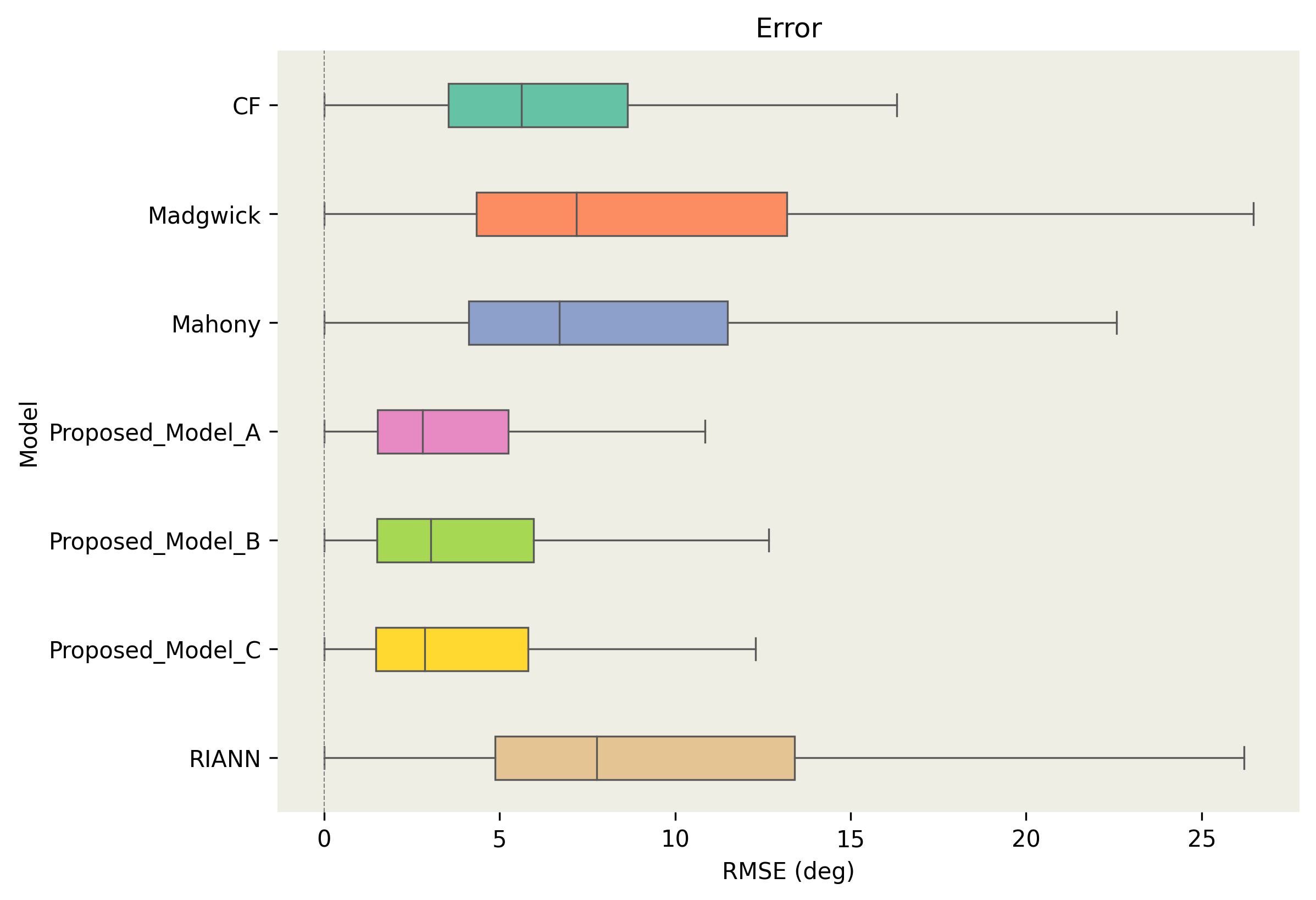}
\caption{OxiOD dataset}
\label{fig:oxiod_boxplot}%
\end{figure}

\begin{figure}[!ht]
\centering
\includegraphics[width=0.45\textwidth]{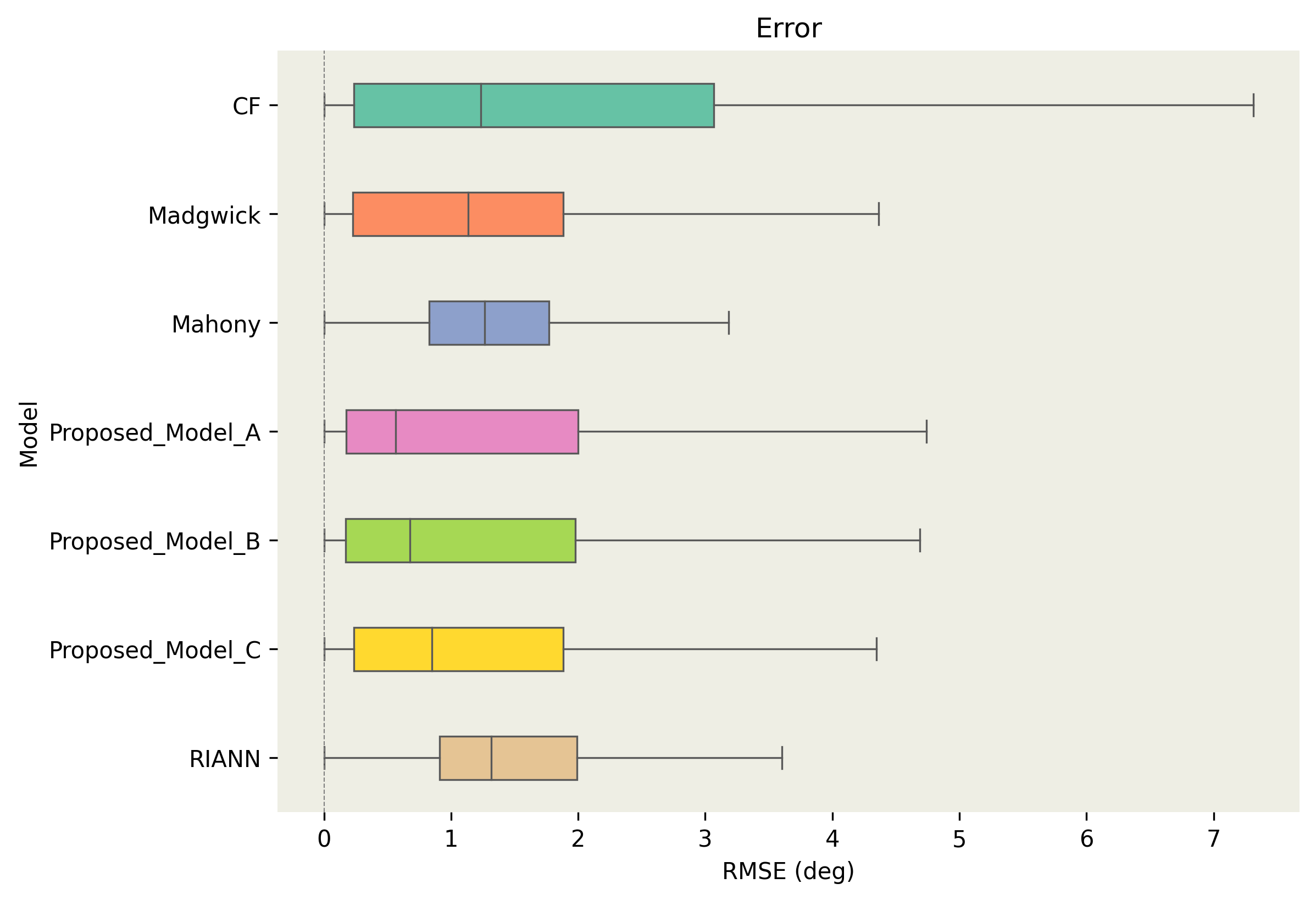}
\caption{Sassari dataset}
\label{fig:sassari_boxplot}%

\end{figure}
\begin{figure}[!ht]%
\centering
\includegraphics[width=0.45\textwidth]{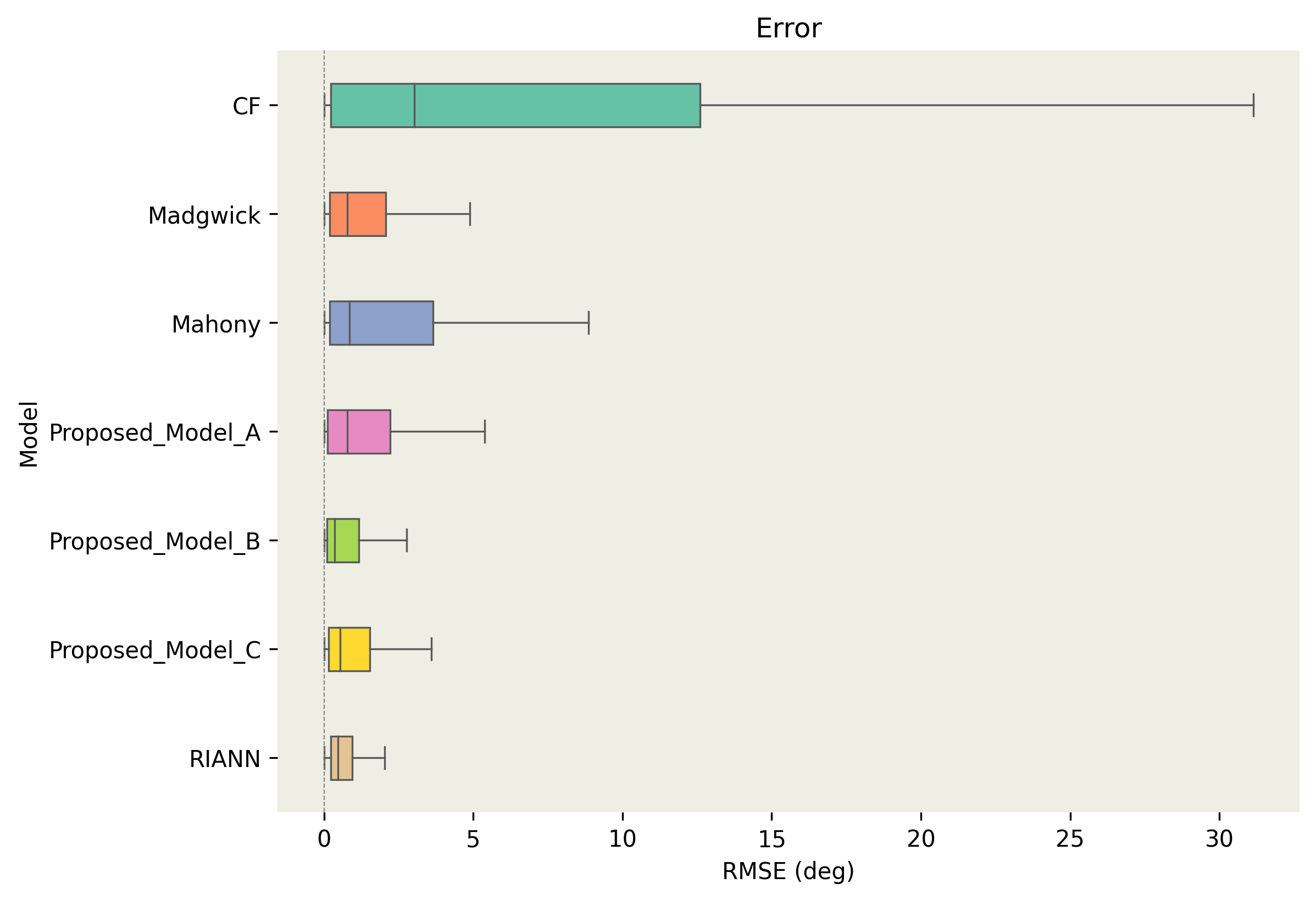}
\caption{Broad dataset}
\label{fig:broad_boxplot}%
\end{figure}

\begin{figure}[!ht]
\centering
\includegraphics[width=0.45\textwidth]{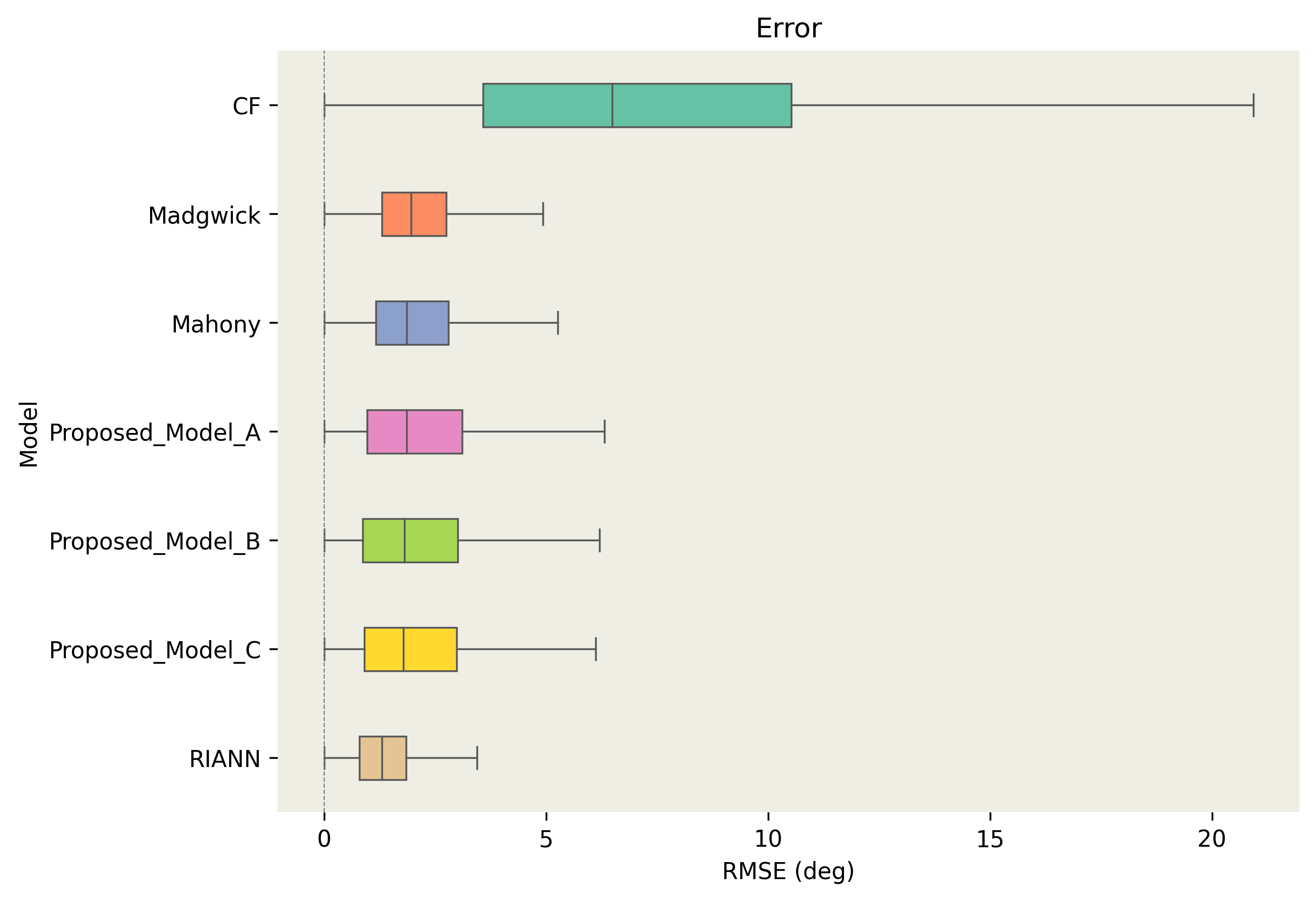}
\caption{RIDI dataset}
\label{fig:ridi_boxplot}%
\end{figure}

\begin{figure}[!ht]
\centering
\includegraphics[width=0.45\textwidth]{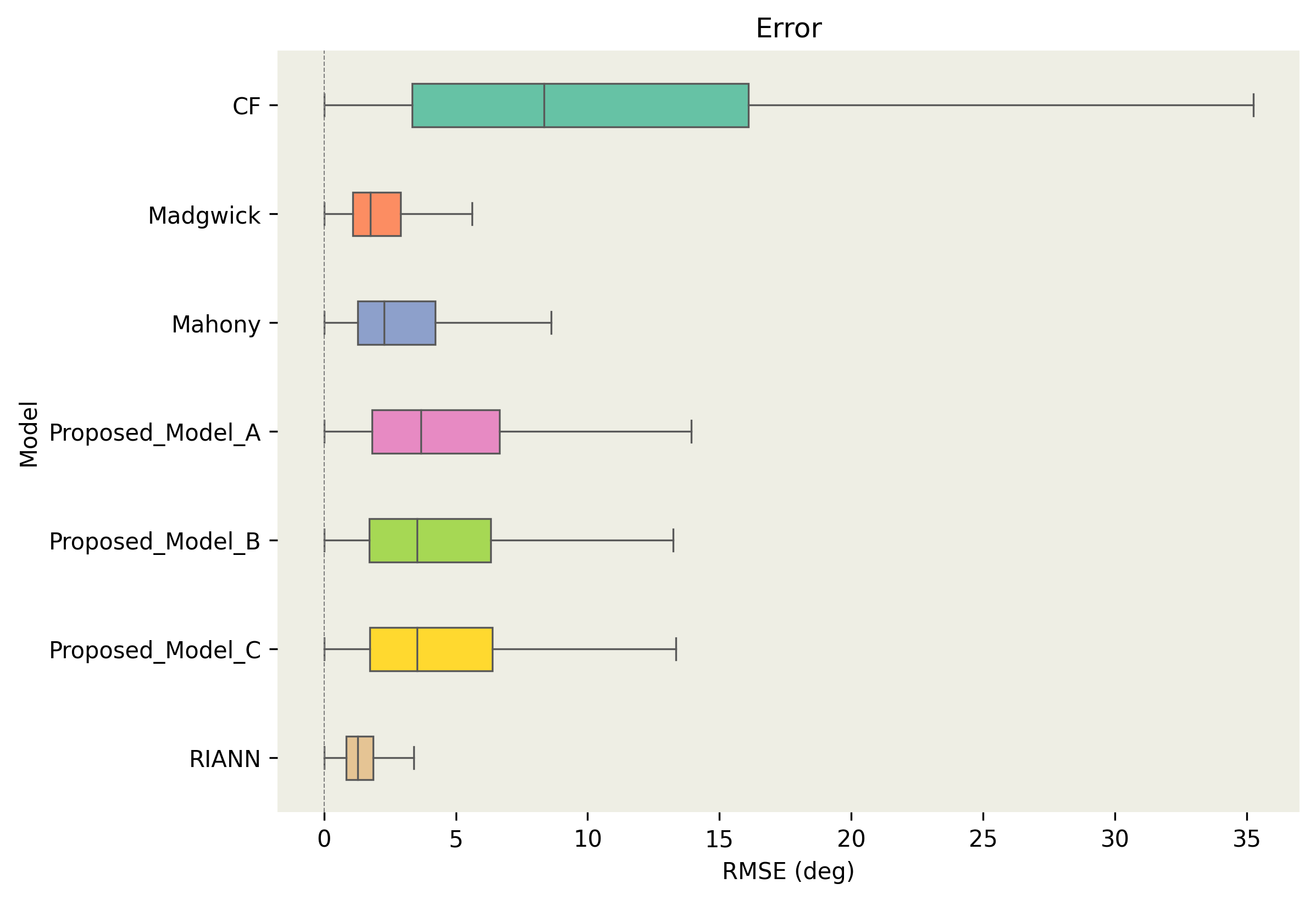}
\caption{RONIN dataset}
\label{fig:ronin_boxplot}
\label{fig:boxplots}
\end{figure}

\section{Results}

The evaluation results on the Sassari dataset demonstrate that the proposed approach, consisting of Model A and Model B, outperformed the other approaches in terms of accuracy and robustness. Specifically, the total rotation error for the proposed method was consistently lower than that of the other methods, with the most significant improvements observed for fast and rotational motions. In contrast, the Madgwick, Mahony, and CF filters exhibited higher total rotation errors, particularly for fast and rotational motion. The RIANN model also exhibited higher total rotation error than the proposed method but performed slightly better than the Madgwick, Mahony, and CF filters.

On the BROAD dataset, the proposed approach, consisting of Model A and Model B, outperformed the other approaches in terms of accuracy and robustness. Specifically, RIANN, Model A, and Model B had the lowest Total Rotation Error values in the majority of the trials, indicating a higher level of accuracy in the attitude estimates. Notably, RIANN used 33 trials of the BROAD dataset to train in three trials for validation. The Madgwick and Mahony filters also performed relatively well, with lower Total Rotation Error values compared to the CF. Additionally, Model A consistently outperformed Model B, suggesting that Model A may be a more effective approach.

Table \ref{table_ridi} includes results for ten different users (94 trials), each corresponding to a different person, and an "Average All" row that shows the average total rotation error across all the dataset. From the table, it is evident that the RIANN Model, performs better than the other approaches in terms of total rotation error. The average total rotation error for Model A is 2.17 degrees, for Model B is 2.03 degrees, and for Model C is 2.06 degrees. The totla error of Model B and Model C is lower than the average total rotation errors of the traditional approaches, which range from 2.13 degrees (Madgwick) to 7.52 degrees (CF).

Table \ref{table_repo} presents the evaluation results of various models on the RepoIMU TStick dataset The dataset consists of 10 different test trials (27 sequences). Each test trial consits of various number of sequences. It was found that the proposed models to be more accurate and robust than other state-of-the-art methods (RIANN, CF, Madgwick, and Mahony) as the MAE, RMSE, and QE values were lower for the proposed method. Specifically, Model A had an average MSE of 5.07, Model B had an average MSE of 3.28 and Model C had an average MSE of 3.36. RIANN had an average MSE of 8.27, CF had an average MSE of 11.98, Madgwick had an average MSE of 11.09, and Mahony had an average MSE of 8.19. Overall \ref{table_repo}, the proposed method appears to be a promising approach for estimating orientation from IMU measurements in the RepoIMU dataset.

Similarly, Table \ref{table_sassari} presents the evaluation results of the proposed methods on the Sassari dataset, demonstrating consistently lower total rotation error than the other methods, particularly for fast and rotational motions. The Madgwick, Mahony, and CF filters had higher total rotation errors, while RIANN performed slightly better than Mahony and CF but worse than the proposed method. Table \ref{table_broad} reports that the proposed method outperforms other approaches on the BROAD dataset, with Model B consistently outperforming Models A and C. However, RIANN used more data for training and validation, while the Madgwick and Mahony filters also had relatively low total rotation error. RIANN used 33 trials of the BROAD dataset to train in 3 trials for the validation.

The RoNIN dataset consists of 152 sequences and 42.7 hours of IMU measurements. So, based on the published dataset, we calculated the mean error over the four main trails (all 152 sequences are subsets of these four main trails). The results in Table \ref{table_ronin} show the evaluation of the proposed method and other approaches on the RoNIN dataset. The proposed method had a close total rotation error compared to other approaches, with more error than RIANN, Madgwick, and Mahony.

Based on the results presented in Table \ref{table_oxiod}, it appears that the proposed Model A performs significantly better than Model B, C and the other approaches (CF, Madgwick, Mahony, and RIANN) in terms of total rotation error on the OxIOD dataset. In almost all 106 trials, Model A had the lowest total rotation error, with an average error of 3.92 degrees. Model B had the second lowest average error at 4.37 degrees, while RIANN had the highest average error at 10.01 degrees. Madgwick had an average error of 9.96 degrees, Mahony had an average error of 8.6 degrees, and CF had an average error of 6.49 degrees.

In Figure~\ref{fig:boxplots}, we present the boxplots illustrating the total rotation error for the proposed method and other approaches. The boxplots display the median, first and third quartiles, and the minimum and maximum values, with whiskers extending to the most extreme data points within 1.5 times the interquartile range from the box. The results indicate that the proposed method consistently outperformed other approaches across diverse motion patterns and sampling rates, with significant improvements noted in fast and rotational motions. Additionally, the proposed method demonstrated robust performance in the presence of environmental disturbances, sensor noise, sampling rate, and motion pattern.

The evaluation results indicate that the proposed methods outperformed conventional filters in terms of accuracy and robustness, with strong generalization capabilities across different motion patterns, sampling rates, and environmental conditions, suggesting it is a promising alternative to conventional attitude estimation filters. Overall, the performance evaluation results demonstrate the effectiveness of the proposed end-to-end deep-learning approach for real-time attitude estimation using inertial sensor measurements. The method offers a high level of accuracy and robustness and shows strong generalization capabilities, making it a promising solution for a wide range of applications.

\section{Limitations and Future Work}

While our attitude estimation model has shown promising results, there are still several limitations that need to be addressed to improve its accuracy and generalization capabilities.

One of the main challenges in using deep learning models for attitude estimation is the availability of large and diverse datasets. Although we utilized multiple datasets to train our model, the lack of diversity in the dataset could still affect its performance on unseen scenarios. To address this issue, we plan to incorporate data augmentation techniques that can help to increase the diversity of the dataset and improve the model's ability to generalize to new scenarios. However, there are currently no proper data augmentation methods available for attitude estimation, which is an area that requires further research.

Moreover, while we employed various hyperparameter optimization methods to select the best model architecture, it is possible that other combinations of hyperparameters could yield better results. However, hyperparameter optimization can be a computationally intensive process, and there may be practical limitations to the extent of hyperparameter tuning that can be performed.

Another limitation of our model is the use of the window of IMU data, which could lead to system delays, especially when the system sampling rate is not high. To address this issue, we plan to investigate the use of other techniques, such as reducing the window, that can reduce the delay in attitude estimation.

Lastly, our model is limited to the use of inertial sensors and does not incorporate magnetometers. Incorporating magnetometers could potentially provide additional information to estimate the yaw angle, which is an important component of the overall attitude estimation. However, this would require additional hardware and software modifications, which is an area that requires further investigation.

In summary, while our attitude estimation model has shown promising results, there are still several limitations that need to be addressed to improve its accuracy and generalization capabilities. These limitations include the availability of large and diverse datasets, hyperparameter optimization, system delays, and the incorporation of magnetometers. Addressing these limitations requires further research and development, and we plan to investigate these areas in future work.

\section{Conclusion}
This study proposes an end-to-end deep learning framework for real-time attitude estimation based on quaternion representation. The proposed models leverage the power of convolutional neural network layers and long short-term memory layers to learn the temporal dependencies between the inertial measurement unit readings and the attitude estimation. Specifically, the CNN layers extract the features from the accelerometer and gyroscope readings, while the LSTM layers are used to capture the temporal relationship between the extracted features.

The proposed models offer several advantages over conventional filters, such as robustness to motion patterns, sampling rates, and environmental disturbances. The models were evaluated on five publicly available datasets, consisting of over 120 hours and 200 kilometers of IMU measurements. The evaluation results demonstrated that the proposed models (Model A, Model B, and Model C) outperformed the state-of-the-art approaches (RIANN, CF, Madgwick, and Mahony) in terms of accuracy and robustness. The proposed methods achieved lower mean absolute error (MAE), root mean squared error (RMSE), and quaternion error (QE) values, indicating a higher level of accuracy in the attitude estimates.

The proposed network architectures consist of four main components: Feature Extraction, Feature Fusion, Sampling Rate Fusion, and Attitude Estimation. The Feature Extraction component extracts the features from the IMU data, and the Feature Fusion component fuses the extracted features. The Sampling Rate Fusion component fuses the sampling rate of the IMU measurements and works as a regularization technique to reduce overfitting. The Attitude Estimation component estimates the attitude using a fully forward neural network with four units followed by a unit scaling layer.

Moreover, the evaluation results demonstrated that the proposed methods have strong generalization capabilities over various motion characteristics and sensor sampling rates. This suggests that the proposed methods can be applied to a wide range of applications without the need for additional optimization or adaptation.

In summary, this study presents an effective and robust deep learning approach for real-time attitude estimation using inertial sensor measurements. The proposed method offers a promising alternative to traditional filters and has the potential to enable a wide range of applications in fields such as robotics, augmented reality, and human-computer interaction. Future research may investigate the integration of other sensors, such as visual or barometric sensors, to further improve the accuracy and robustness of the attitude estimation. Additionally, extending the proposed method to related tasks, such as orientation tracking or pose estimation, is another avenue for future research.

\section*{Aacknowledgements}
The study presented in this paper is based on A. Asgharpoor Golroudbari's M.Sc. Thesis ("Design and Simulation of Attitude and Heading Estimation Algorithm", Department of Aerospace, Faculty of New Sciences \& Technologies, University of Tehran).

We would like to express our sincere gratitude to Prof. Parvin Pasalar, Dr. Farsad Nourizade, and Dr. Maryam Karbasi Motlagh at the Students' Scientific Research Center at Tehran University of Medical Sciences. Their invaluable scientific advice and support in the form of access to computational resources were instrumental in the success of our research. We are deeply appreciative of their contributions and the time they dedicated to helping us.
\section*{Conflict of Interest}
The authors declare that they have no known competing financial interests or personal relationships that could have appeared to influence the work reported in this paper.

\section*{Data availability}
The code of the proposed models has been made publicly available \cite{Asgharpoor_Golroudbari_End-to-End-Deep-Learning-Framework-for-Real-Time-Inertial-Attitude-Estimation-using-6DoF-IMU_2023}.

\section*{Appendix}
\appendix

\section{IMU Dynamics Model}
In this section, we discuss the dynamics model of an IMU, which is composed of a gyroscope and an accelerometer that measure the angular velocity and linear acceleration of a rigid body. The measurements from the IMU are subject to noise and bias, where the noise is a random process that is independent of the previous measurements, and the bias is a systematic error that is constant or slow-varying over time. To describe the relationship between the measurements and the orientation of the rigid body, we use a nonlinear IMU dynamics model. The model is represented by equations \ref{eq:gyro_model} and \ref{eq:acc_model}, where $\omega$ is the angular velocity of the rigid body, $\mathbf{a}$ is the linear acceleration of the rigid body, $\mathbf{R}$ is the rotation matrix that describes the orientation of the body frame with respect to the inertial frame, $\mathbf{g}$ is the gravity vector, and $\mathbf{b}{\omega}$ and $\mathbf{b}{a}$ are the biases of the gyroscope and accelerometer, respectively. Additionally, $\mathbf{v}{\omega}$ and $\mathbf{v}{a}$ are the noise of the gyroscope and accelerometer, respectively.
\begin{equation}
\mathbf{v}_{\omega} \sim \mathcal{N}(0, \sigma_{\omega}^2) \label{eq:gyro_noise}
\end{equation}
\begin{equation}
\mathbf{v}_{a} \sim \mathcal{N}(0, \sigma_{a}^2) \label{eq:acc_noise}
\end{equation}
\begin{equation}
\tilde{\omega} = \omega - \mathbf{b}_{\omega} + \mathbf{v}_{\omega} \label{eq:gyro_model}
\end{equation}	
\begin{equation}
\tilde{\mathbf{a}} = (\mathbf{R}^T \mathbf{g}) + \mathbf{a} - \mathbf{b}_{a} + \mathbf{v}_{a} \label{eq:acc_model}
\end{equation}	
The IMU dynamics model is crucial for accurately estimating the attitude of the rigid body in real-time. The model's nonlinear nature poses a challenge in accurately estimating the attitude due to the accumulated errors from the noise and bias. Therefore, it is important to consider these factors when developing an attitude estimation algorithm for IMUs. The equations presented in this section provide the foundation for developing such algorithms.
\section{Orientation}
In the context of rigid body motion, the orientation of an object with respect to a reference frame can be defined as the shortest rotation between a frame attached to the object and the reference frame. The orientation parameters, also known as attitude coordinates, represent a set of parameters that fully describe the attitude of a rigid body. However, there are multiple ways to represent the attitude of a rigid body, and the choice of representation depends on the specific application requirements.

The most common attitude representations are Euler angles, rotation matrices, and quaternions. Euler angles define rotations about the three orthogonal axes of the body frame, which are typically referred to as yaw, pitch, and roll (or heading, elevation, and bank). However, Euler angles suffer from the gimbal lock problem, which limits their applicability in certain cases. Rotation matrices, on the other hand, are 3x3 matrices that represent the orientation of the body frame with respect to the inertial frame. However, this representation leads to six redundant parameters, which can result in numerical instability and computational complexity.

Quaternions are a 4x1 vector that offers an alternative representation of attitude that is more suitable for attitude estimation. Quaternions are not subject to the gimbal lock problem and have the least redundant parameters, making them more numerically stable and computationally efficient. To avoid singularities and minimize redundant parameters, we use quaternion representation with the components $[w, x, y, z]$, rather than Direction Cosine Matrix (DCM) or Euler angles. The quaternion can be defined as follows:

\begin{equation}
\begin{gathered}
\mathbf{q} =
\begin{bmatrix}
q_0 &
q_1 &
q_2 &
q_3
\end{bmatrix}^T
\end{gathered}
\end{equation}

Here, $q_0$ is the scalar part and $q_1$, $q_2$, and $q_3$ are the vector part. The relationship between quaternions and Euler angles is given by the following equation:

\begin{equation}
\begin{gathered}
\mathbf{q} =
\begin{bmatrix}
C(\phi/2) C(\theta/2) C(\psi/2) + S(\phi/2) S(\theta/2) S(\psi/2) \\
S(\phi/2) C(\theta/2) C(\psi/2) - C(\phi/2) S(\theta/2) S(\psi/2) \\
C(\phi/2) S(\theta/2) C(\psi/2) + S(\phi/2) C(\theta/2) S(\psi/2) \\
C(\phi/2) C(\theta/2) S(\psi/2) - S(\phi/2) S(\theta/2) C(\psi/2)
\end{bmatrix}
\end{gathered}
\end{equation}
  
Note that C and S are shorthand for cosine and sine, respectively.

Here, $\phi$, $\theta$, and $\psi$ are the Euler angles. To avoid singularities and have the least number of redundant parameters, the quaternion representation with components $[w, x, y, z]$ is preferred over the Direction Cosine Matrix (DCM) or Euler angles.
The disparity between the estimated and actual attitude can be computed by means of the quaternion multiplicative error, which is expressed by the following equation:
\begin{equation}
\begin{gathered}
\mathbf{q}_{err} = \mathbf{q}_{true} \otimes \mathbf{q}_{est}^{-1}
\end{gathered}
\end{equation}
Here, $\mathbf{q}{err}$ denotes the shortest rotation between the true and estimated orientation, and the operator for quaternion multiplication is defined as follows:
\begin{equation}
\begin{gathered}
\mathbf{q} \otimes \mathbf{p} = \begin{bmatrix}
q_0p_0 - q_1p_1 - q_2p_2 - q_3p_3 \\
q_0p_1 + q_1p_0 + q_2p_3 - q_3p_2 \\
q_0p_2 - q_1p_3 + q_2p_0 + q_3p_1 \\
q_0p_3 + q_1p_2 - q_2p_1 + q_3p_0
\end{bmatrix}
\end{gathered}
\end{equation}
In this equation, $\mathbf{q}$ and $\mathbf{p}$ are the quaternions to be multiplied, and the angle between the actual and estimated orientation can be determined by the following expression:
\begin{equation}
\begin{gathered}
\theta = 2 \arccos( \text{scalar}( \mathbf{q}_{err}) )
\end{gathered}
\end{equation}
Here, $\theta$ is the angle between the true and estimated orientation. Using these equations, the cumulative error of the estimated attitude over $N$ time steps can be obtained as:
\begin{equation}
\begin{gathered}
e_{\alpha}= 2 \arccos( \text{scalar}( \mathbf{q}_{err}) )
\end{gathered}
\end{equation}
\begin{equation}
\begin{gathered}
RMSE = \sqrt{\frac{1}{N} \sum{i=1}^{N} e_{\alpha}^2}
\end{gathered}
\end{equation}
Here, $N$ denotes the number of samples, and $e_{\alpha}$ represents the angle between the true and estimated orientation. The RMSE denotes the root mean square error of the estimated attitude, which indicates the difference between the values predicted by a model or estimator and the values observed. A lower RMSE indicates a better fit of the model to the data.

\section{Sequential Modeling}

Time series data comprises observations taken at regular intervals over time and can be employed for various applications such as forecasting, anomaly detection, and classification. IMU measurements can also be treated as time series data. Time series estimation entails predicting future values in a time series based on past observations, and can be employed as a technique to forecast the future orientation of IMU sensors. Sequential modeling is a method to achieve time series estimation, wherein a deep learning model can learn the relationship between input and output data.

Sequence models can handle sensor data, text, sound, and data with an underlying sequential structure for several applications including time series data prediction \cite{rajagukguk2020review}, speech recognition \cite{graves2013speech}, natural language processing \cite{schoene2022narrative}, music generation \cite{marinescu2019bach}, and DNA sequence analysis \cite{shen2018recurrent}. Traditional neural network models cannot handle time-series data since they do not loop and fail to address time dependencies between them. Recurrent Neural Networks, Long Short-Term Memory networks, Gated Recurrent Units (GRUs) , and Temporal Convolutional Networks are common models used for sequential modeling. RNNs can capture long-term dependencies in data but can be slow due to back-propagation through time. LSTMs employ memory cells to store information from previous inputs. GRUs use gating mechanisms to control the flow of information, while TCNs use dilated causal convolutions, allowing them to learn patterns over longer sequences while maintaining the computational efficiency of traditional Convolutional Neural Networks (CNNs). Each model has its strengths and weaknesses, and the optimal choice depends on the specific task at hand.

In summary, sequential modeling is an effective method for time series estimation, which can be used for inertial attitude estimation. Various deep learning models, including RNNs, LSTMs, GRUs, and TCNs, can be used for sequential modeling. The choice of model depends on the specific task requirements, such as computational efficiency, long-term dependencies, or the ability to learn patterns over longer sequences.

Deep learning models offer an efficient solution to handle sequential data such as measurements from IMUs by uncovering the underlying relationships between input and output data. However, there are several neural network architectures available, each with distinct characteristics and advantages.

For instance, the feedforward neural network is a simple model suitable for classification problems. On the other hand, the convolutional neural network is adept at signal processing and extracting features from input data. Despite this, it does not possess the ability to store data from previous time steps.

To overcome this limitation, the recurrent neural network was developed. RNN is a type of neural network that can store data from previous time steps by using memory cells. Gated recurrent units are a type of RNN that employs gating mechanisms to control the flow of information. Specifically, GRUs have two gates, the reset gate, which controls the flow of information from previous time steps, and the update gate, which controls the flow of information from the current time step.

Another type of RNN that is capable of learning long-term dependencies is the long short-term memory. LSTM has three gates: the input gate, which controls the information that enters the cell state; the forget gate, which controls the information that leaves the cell state, and the output gate, which controls the information that is output.

Temporal convolutional network is another type of neural network that is suitable for sequential data. TCN is a stack of dilated convolutional layers with residual connections that extract features from the input data. The residual connections preserve information from previous time steps, allowing TCN to capture temporal dependencies efficiently.

It is essential to choose the appropriate model based on the specific task at hand, as each model has its strengths and weaknesses. In the following sections, we will delve into these models in greater detail, including their architecture, training, and applications.

\subsection{Convolutional Neural Network}
A Convolutional Neural Network is an artificial neural network that is specifically designed for analyzing data with a spatial or temporal structure. The CNN achieves this through the use of convolutions, which are mathematical operations that enable the network to extract features from the input data. These networks have gained wide-spread popularity in image processing, computer vision, and natural language processing. Various filters are utilized in a convolutional layer to detect specific features in the input data. The output of these filters is then passed through a nonlinear activation function, such as ReLU or sigmoid, before being fed to another layer of the network. This process enables object recognition, segmentation, and classification, by learning patterns from images or videos through multiple layers with different parameters and associated weights. Recent advances in deep learning have demonstrated that CNNs can also be utilized for time-series prediction \cite{li2022stock,chiang2021hybrid} and nonlinear regression \cite{fan2022soc}.

The CNN is composed of two primary types of layers, convolutional layers and pooling layers. The convolutional layers are responsible for extracting features from the input data. The architecture of a CNN is depicted in figure~\ref{fig1}. The convolutional layer is defined by the following equation \cite{koushik2016understanding}:

\begin{figure}[!ht]
	\centering \includegraphics[width=0.5\linewidth]{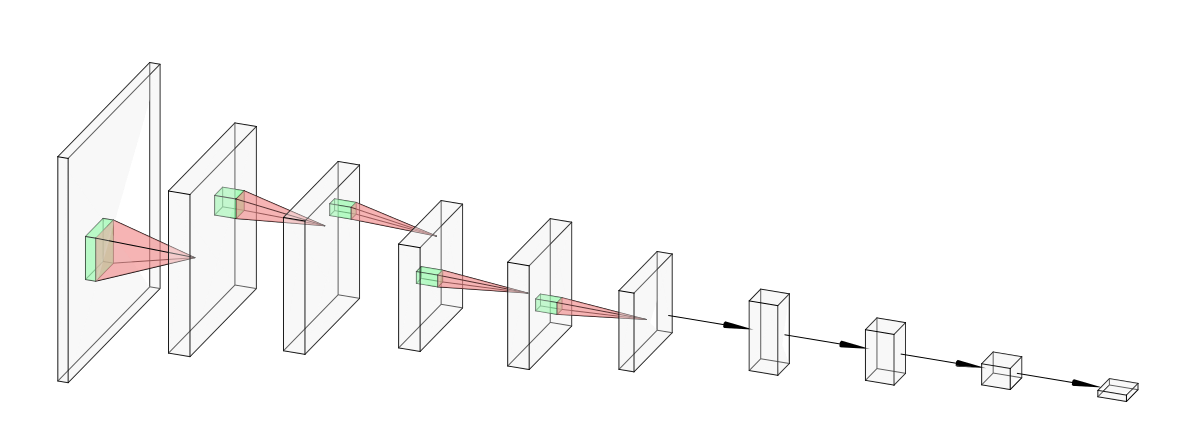}
	\caption{Convolutional Neural Network. {\footnotesize{This image was created at \href{http://alexlenail.me/NN-SVG/AlexNet.html}{http://alexlenail.me/NN-SVG/AlexNet.html}}}}\label{fig1}
\end{figure}
\begin{equation}
	\begin{gathered}
		y_k \equiv \sum_{i=k}^{k+W-1} x_i w_{k+W-i}
	\end{gathered}
\end{equation}

where $x$ represents the input data, and $w$ is the filter size. Following the convolutional layer, a pooling layer is typically employed to extract the most significant features from the input data while reducing its dimensionality. The pooling layer consists of a pooling function and a pooling window, as shown in figure~\ref{pooling}. 

\begin{figure}[!ht]
	\centering \includegraphics[width=0.5\linewidth]{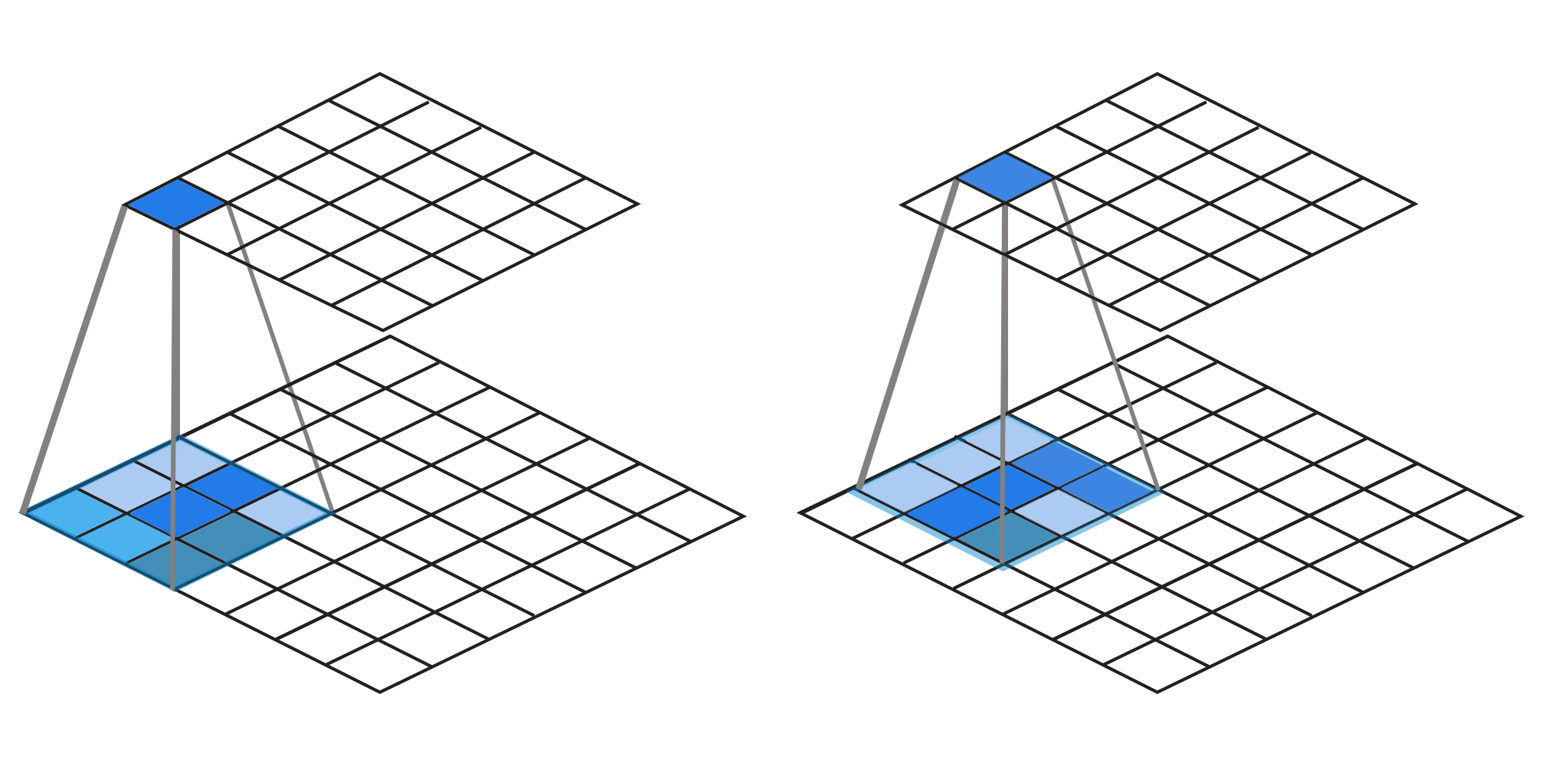}
	\caption{Pooling Layer.}\label{pooling}
\end{figure}

 Several types of pooling layers are available, including max pooling, average pooling, and global average pooling. Max pooling takes the maximum value from each region of an input feature map, while average pooling takes the mean value from each region. Global average pooling reduces a feature map to a single number by taking the mean across all regions in the input feature map \cite{he2015spatial}. 

Through the use of convolutional and down-sampling techniques, CNNs transform the original input layer by layer, producing class scores for classification and regression.

\subsection{Recurrent Neural Network}

\begin{figure}[!ht]
	\centering \includegraphics[width=3.0in]{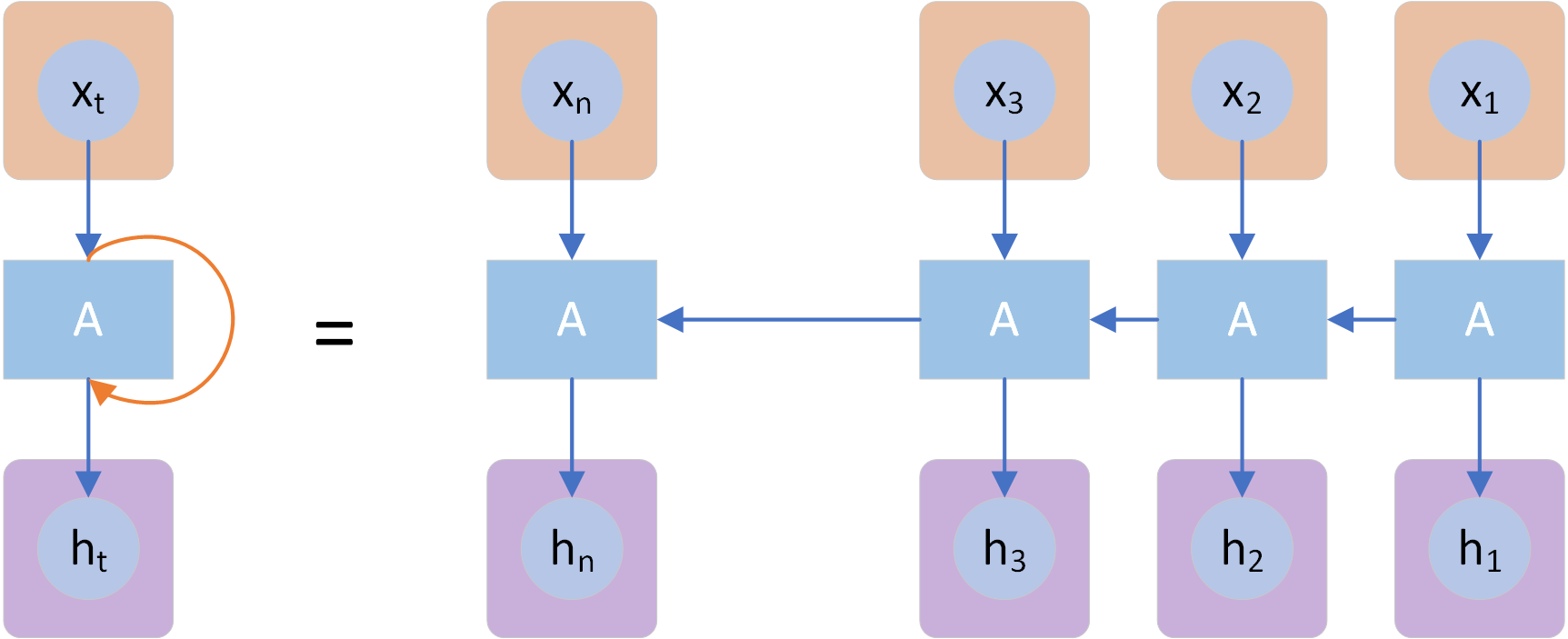}
	\caption{Recurrent Neural Network}\label{fig3}
\end{figure}

A Recurrent Neural Network is a type of artificial neural network designed for sequential data processing. RNNs have the ability to store and remember previous inputs, allowing them to process sequential data like text and audio. The network comprises several layers of neurons that are interconnected in a cyclic manner, enabling the output from one layer to serve as input for another layer and vice versa. By passing information across time steps through multiple layers, RNNs can capture long-term dependencies between elements in a sequence. These networks can be utilized for tasks such as speech recognition and language translation by learning patterns in the input sequence over time through different parameters and weights assigned to each neuron in each layer. Figure~\ref{fig3} shows the architecture of an RNN.

The RNN equation is given by:
\begin{equation}
\begin{gathered}
h_t = \sigma(W_{hh}h_{t-1} + W_{xh}x_t + b_h)
\end{gathered}
\end{equation}
where $h$ represents the hidden state, $x$ represents the input data, $W_{hh}$ denotes the weight matrix for the hidden state, $W_{xh}$ denotes the weight matrix for the input data, and $b$ denotes the bias term. The sigmoid function $\sigma$ is used to ensure that the hidden state output remains within a certain range. RNNs are widely used in various applications such as speech recognition, language modeling, and machine translation. They are effective in processing sequences of variable length and handling sequential data with long-term dependencies.
\subsection{Long Short-Term Memory}

LSTM is a type of RNN that utilizes memory cells to store and retrieve information from previous inputs. This allows the model to recognize and learn patterns over long sequences of data, making it a powerful time domain deep learning model \cite{cai2021short}. The architecture of LSTM is composed of three fundamental components: an input gate, a forget gate, and an output gate. The input gate controls which values are added to the cell state; the forget gate controls which values are removed from it, and the output gate determines what is passed out as output for each time step in sequence processing tasks, such as machine translation or speech recognition. The key advantage of LSTMs is their ability to capture long-term dependencies in the input data.

\begin{figure}[!ht]
\centering
\includegraphics[width=.45\textwidth]{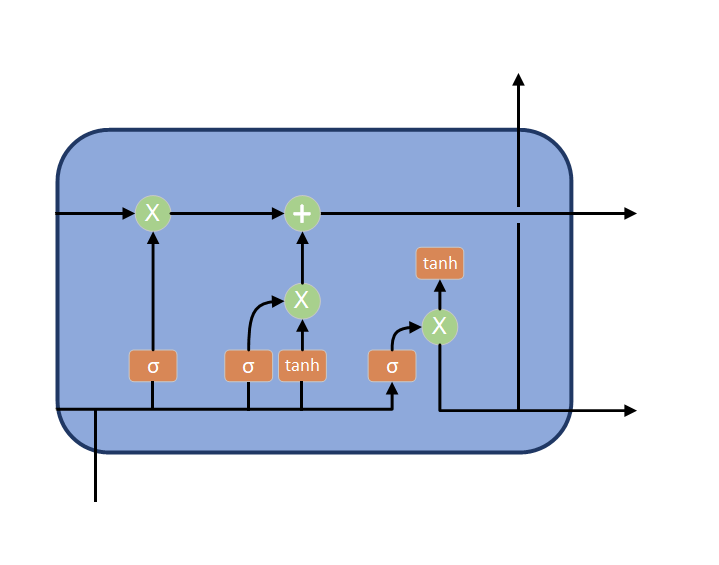}
\caption{Long Short-Term Memory}\label{fig4}%
\end{figure}

The LSTM architecture is depicted in Figure~\ref{fig4}, and its equations are as follows \cite{hochreiter1997long}:
\begin{align}
i_t &= \sigma(W_i \cdot [h_{t-1}, x_t] + b_i) \\
f_t &= \sigma(W_f \cdot [h_{t-1}, x_t] + b_f) \\
o_t &= \sigma(W_o \cdot [h_{t-1}, x_t] + b_o) \\
c_t &= f_t \cdot c_{t-1} + i_t \cdot \mathrm{tanh}(W_c \cdot [h_{t-1}, x_t] + b_c)
\end{align}

Here, $i_t$, $f_t$, and $o_t$ denote the input, forget, and output gates, respectively. $c_t$ is the cell state at time $t$. $h_{t-1}$ and $x_t$ are the hidden state and input data, respectively. $W_i$, $W_f$, $W_o$, and $W_c$ are the weight matrices for the input, forget, output, and cell state update equations, respectively, and $b_i$, $b_f$, $b_o$, and $b_c$ are the bias terms. $\mathrm{tanh}$ is the hyperbolic tangent activation function.

\subsection{Gated Recurrent Unit}

\begin{figure}[!ht]
\centering
\includegraphics[width=.45\textwidth]{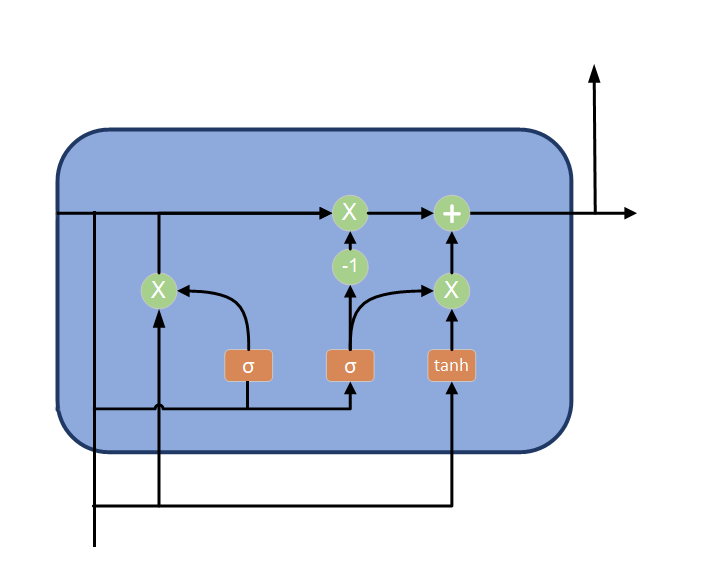}
\caption{Gated Recurrent Unit}\label{fig5}%
\end{figure}

GRU is a variant of the recurrent neural network architecture that bears resemblance to the Long Short-Term Memory network. GRUs use gating mechanisms that regulate the flow of information within the network, enabling it to more effectively capture long-term dependencies within sequential data. In contrast to LSTMs, GRUs have fewer parameters, making them computationally less expensive and easier to train. The architecture of the GRU is depicted in fig.~\ref{fig5}. Its equations are given as \cite{cho2014properties}:
Update Gate:
\begin{equation}
\begin{gathered}
z_t = \sigma(W_z \cdot [h_{t-1}, x_t] + b_z)
\end{gathered}
\end{equation}
Reset Gate:
\begin{equation}
\begin{gathered}
r_t = \sigma(W_r \cdot [h_{t-1}, x_t] + b_r)
\end{gathered}
\end{equation}
Hidden State Update :
\begin{equation}
\begin{gathered}
h_t = z_t \cdot h_{t-1} + (1 - z_t) \cdot W \cdot [r_t, x_t] + b
\end{gathered}
\end{equation}


LSTMs and GRUs share the same objective of preserving long-term memory in sequential data. In LSTMs, memory cells store information from past inputs, facilitating pattern recognition across lengthy sequences and enabling accurate predictions based on such patterns. The input gate controls the selective addition of values to the memory cell, the forget gate controls the selective removal of values from it, and the output gate regulates what output is generated at each step of sequence processing tasks, such as machine translation or speech recognition. On the other hand, GRUs use update and reset gates to control the flow of information within the network. The update gate regulates how much of a new value should be stored in the hidden state while resetting any irrelevant previously stored information, thereby enabling GRUs to learn long-term dependencies more effectively than conventional RNNs without compromising speed or accuracy. The choice between the two models depends on the specifics of the task at hand, as the performance of one may be superior to the other in some instances, while they may perform equally well in others \cite{gao2020short, wang2020real, gruber2020gru}. Both models find applications in fields such as speech recognition or machine translation.

\subsection{Temporal Convolutional Network}

TCNs are a type of neural network that has been developed for sequence modeling tasks, such as machine translation, speech recognition, and time series forecasting. TCNs achieve this by using dilated causal convolutions to capture long-term dependencies in data while maintaining the computational efficiency of traditional CNNs.

The TCN model is made up of multiple layers with increasing dilation factors. Each layer is composed of a set of convolutional filters with different kernel sizes and numbers of filters. The input data is passed through each convolutional layer, where the filters apply temporal convolutions to the input data. The kernel size determines the number of time steps that the convolutional filters are applied to, allowing the model to capture longer-term dependencies in the input data. The number of filters determines the number of output feature maps generated by the convolutional layers.

\begin{figure}[!ht]
	\centering \includegraphics[width=\linewidth]{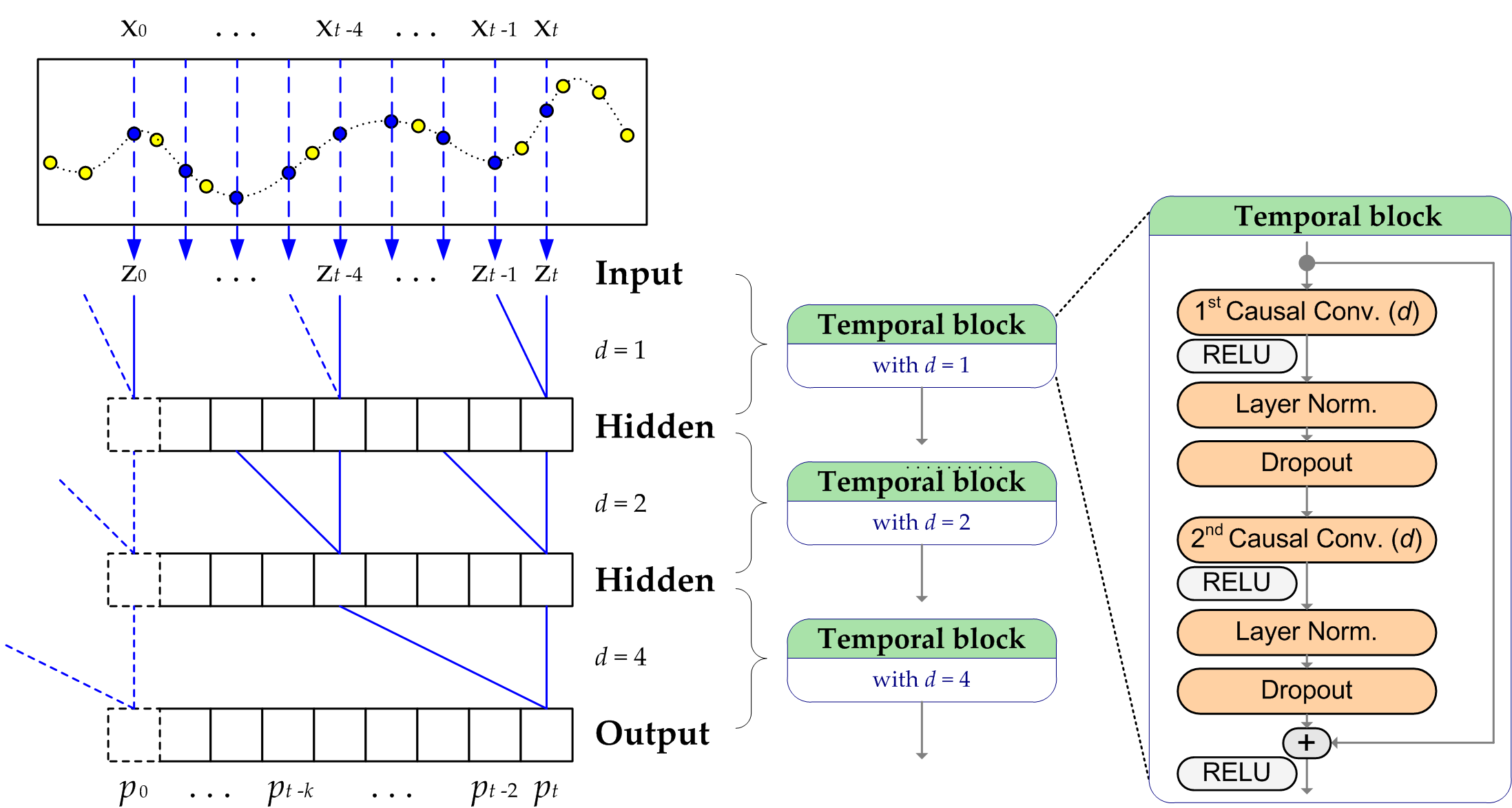}
	\caption{Temporal Convolutional Network \cite{moor2019early}}\label{fig6}
\end{figure}

The TCN architecture is depicted in Figure \ref{fig6}, and the TCN equation is represented as follows \cite{lea2017temporal}:

\begin{equation}
\begin{gathered}
y_t = f(W*x_t + b),
\end{gathered}
\end{equation}

where $W$ is the weight matrix, $x_t$ is the input at time $t$, and $b$ is the bias vector. By applying dilated convolutions to the input data, TCNs can increase the receptive field of the filters without increasing the number of parameters, which enables the model to capture longer-range dependencies in the input data without increasing the computational complexity.

TCNs are highly effective in handling sequential data and have been successfully applied to a wide range of tasks such as natural language processing, time series forecasting, and speech recognition. Compared to other models, TCNs are computationally efficient and have been shown to achieve high accuracy with smaller numbers of parameters, making them particularly useful for real-time applications.

\subsection{Bidirectional Layer}
\begin{figure}[!ht]
	\centering \includegraphics[width=0.5\linewidth]{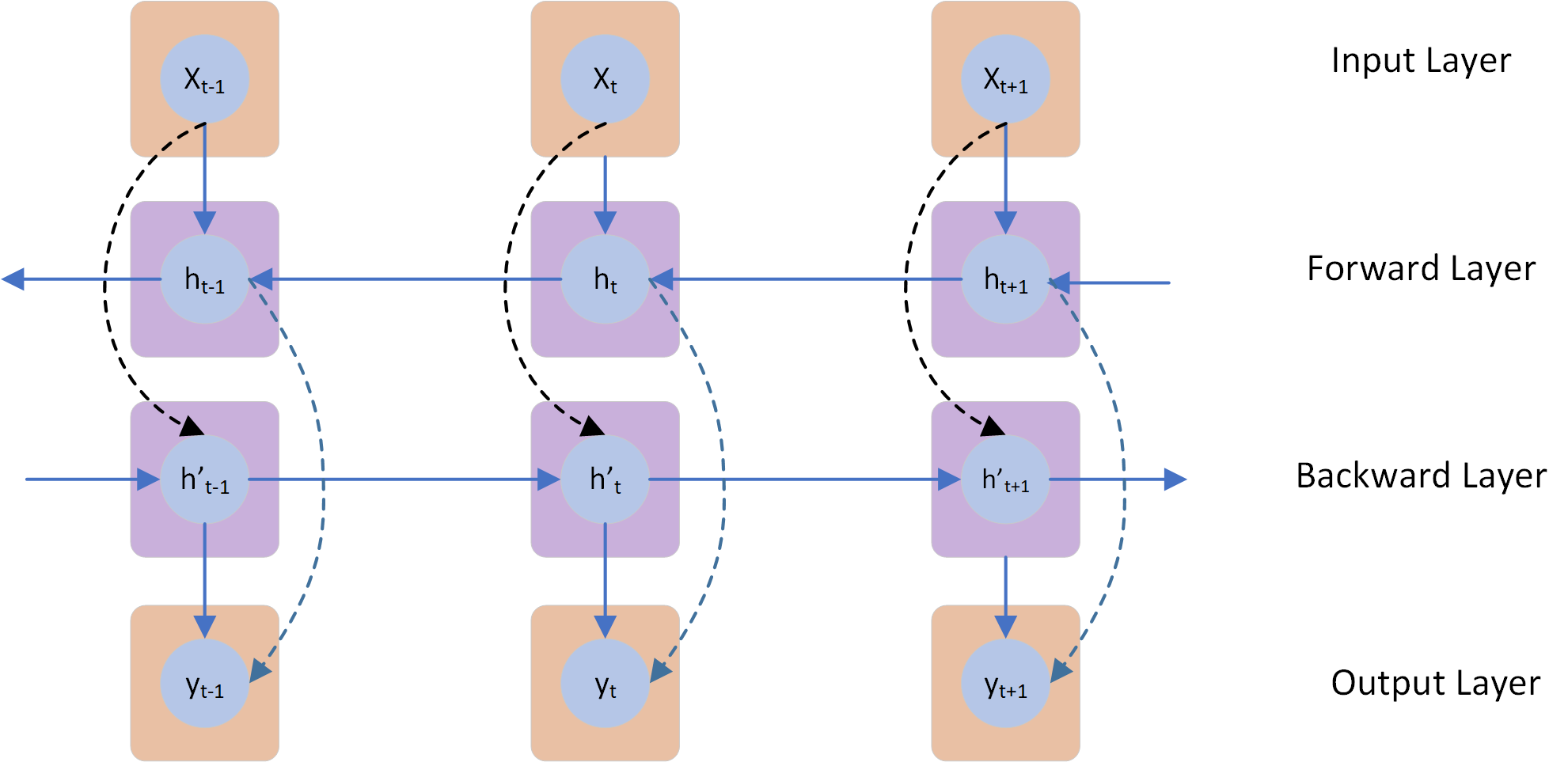}
	\caption{Bidirectional Layer}\label{fig7}
\end{figure}

A bidirectional layer is a neural network layer that processes data bidirectionally, allowing the model to incorporate information from both past and future observations to overcome the problem of accumulative errors. The bidirectional layer was first introduced by Schuster and Paliwal in 1997 \cite{schuster1997bidirectional} and has since become a widely used component in many neural network architectures. The bidirectional layer consists of two separate layers, each of which processes data in opposite directions (forward and backward) to produce an output vector. These two output vectors are then concatenated to form a single output vector that can be used for prediction or classification tasks. The architecture of the bidirectional layer is shown in Figure~\ref{fig7}. The bidirectional layer is typically used in RNN architectures, such as LSTM networks, to improve their performance in tasks involving sequential data, such as speech recognition and natural language processing. The bidirectional layer has also been successfully applied in other types of neural networks, such as CNNs, to capture both local and global features in images.

\section{Activation Function}

Activation functions are critical components of neural networks that introduce non-linearity into the models. They transform the output of a neuron into a format that can be further processed, enabling the network to learn complex patterns and represent data better. The activation function determines how the neural network should respond to specific inputs, and it helps control the flow of information, enabling it to make decisions based on certain inputs.

Activation functions have different properties that can be useful for different tasks, such as classification or regression problems. For instance, Piecewise Linear Activation Functions are composed of a limited number of linear segments, each defined over an equal number of intervals. They are commonly used in Artificial Neural Networks to provide the necessary non-linearity for the model to learn complex representations. Rectified Linear Unit (ReLU) is an example of Piecewise Linear Activation Function, which has a constant first-order derivative and no curvature in each interval defined by its breakpoint. On the other hand, Locally Quadratic Activation Functions are non-linear, smooth activation functions with nonzero second derivatives that can be approximated by a quadratic equation in a specific area. These functions are locally quadratic, meaning they can be represented by a parabola in a certain region, but may not be a perfect parabola everywhere.

\begin{figure*}[h!]
\centering
\includegraphics[width=0.9\textwidth]{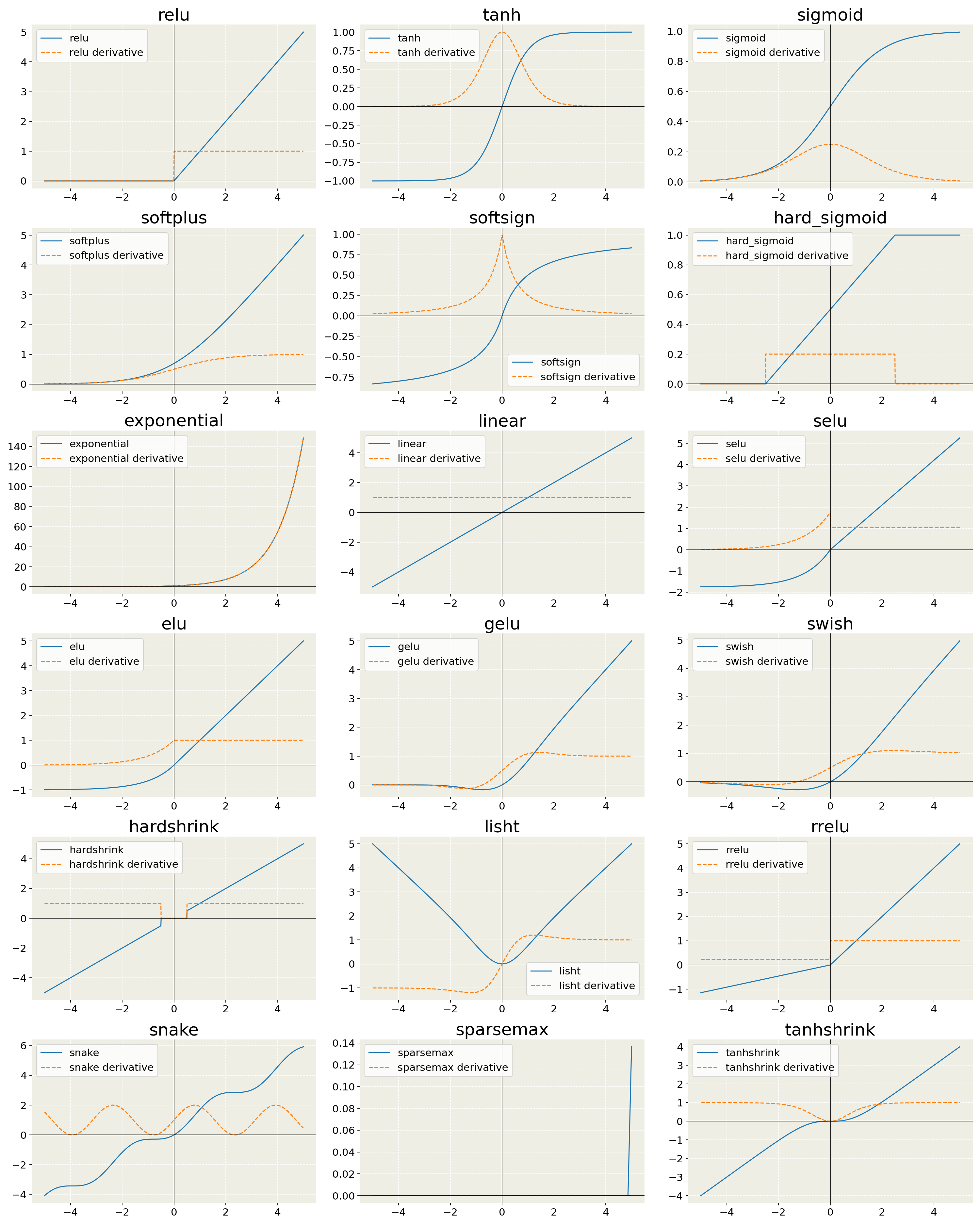}
\caption{Piecewise Linear and Locally Quadratic Activation Functions}\label{activation_functions}
\end{figure*}

\begin{figure}[!ht]
\centering
\centering
\includegraphics[width=0.5\linewidth]{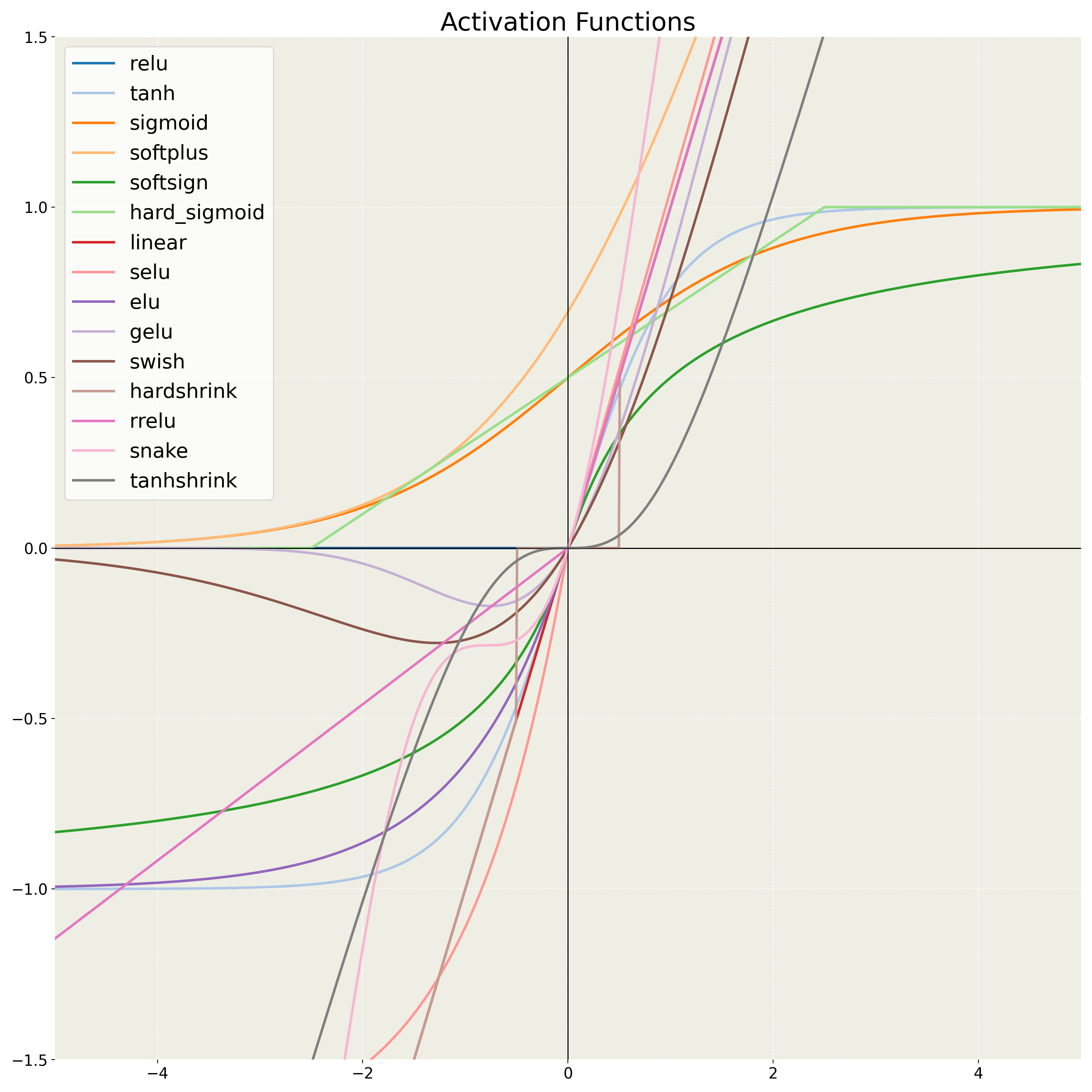}
\caption{Performance Comparison of Different Activation Functions \cite{golroudbari2023recent}}\label{activation_function_compare}
\end{figure}

Figures~\ref{activation_functions} and \ref{activation_function_compare} show the most common activation functions in neural networks. The former displays Piecewise Linear and Locally Quadratic Activation Functions, while the latter compares the performance of different activation functionns. In summary, activation functions play a vital role in enabling neural networks to learn complex patterns and represent data effectively, and selecting the right activation function for a specific task can significantly impact the model's performance.

\subsection{Sigmoid Function}

The sigmoid function is maps the input $x$ from the domain of real numbers $(-\infty,\infty)$ to the range of probabilities between 0 and 1, which are interpreted as the likelihood that the input belongs to a certain class. The sigmoid function can be defined as follows:

\begin{equation}
f(x) = \frac{1}{1+e^{-x}}
\end{equation}

where $e$ is the Euler's number and $x$ is the independent variable. The output of the sigmoid function ranges from 0 to 1, with values close to zero indicating low probabilities and values close to one indicating high probabilities. The sigmoid function has been extensively used in binary classification tasks but can also be applied to multi-class problems.

\subsection{Rectified Linear Unit}

Rectified Linear Unit (ReLU) is a widely used activation function in neural networks. It transforms an input $x$ to an output that is equal to $x$ when $x$ is positive and 0 otherwise. The ReLU function can be mathematically expressed as:

\begin{equation}
f(x) = max(0,x)
\end{equation}

where $x$ is the independent variable. The output of the ReLU function ranges from 0 to infinity, with values close to zero indicating low probabilities and higher values indicating higher probabilities. ReLU is particularly suitable for image recognition tasks since it allows for faster training times compared to other activation functions like sigmoid or hyperbolic tangent (tanh). However, it is not well suited for tasks requiring negative values such as regression problems.

\subsection{Hyperbolic Tangent}

The Hyperbolic Tangent is an activation function introduced by Sepp Hochreiter \cite{hochreiter1997long}. It maps the input $x$ to an output between -1 and 1, making it useful for classification tasks. The tanh function can be used as an alternative to the sigmoid activation function for training deep neural networks and helps to mitigate the vanishing gradient problem associated with other activation functions such as ReLU or ELU. The tanh function is defined as:

\begin{equation}
f(x) = \frac{e^x-e^{-x}}{e^x+e^{-x}}
\end{equation}

\subsection{Leaky ReLU}

The Leaky ReLU is an activation function that overcomes the limitation of the regular ReLU function by allowing for negative input values. It produces a small fraction of negative input values, known as the "leak," while returning the input for positive values. The Leaky ReLU can be mathematically defined as:

\begin{equation}
f(x) = max(0,x) + \alpha \min(0,x)
\end{equation}

where $x$ is the independent variable, $\alpha$ is the leak parameter (usually set to 0.01), and the output of the equation ranges from $-\alpha$ to infinity. The Leaky ReLU activation function is commonly used for image recognition tasks, as it provides good accuracy results while still allowing for faster training times than other activation functions such as sigmoid or tanh.

\subsection{Exponential Linear Unit}

The Exponential Linear Unit (ELU) is an activation function proposed by Djork-Arné Clevert in 2015 \cite{heusel2015elu}. It has shown to be effective for deep neural network training. The ELU activation function is similar to the ReLU, but it has a negative part, which enables it to mitigate the vanishing gradient problem associated with other activation functions, such as sigmoid or tanh.

The ELU is formulated as:

\begin{equation}
\begin{gathered}
f(x) = \begin{cases}
x & \text{if } x \geq 0 \\
\alpha(e^x-1) & \text{if } x < 0
\end{cases}
\end{gathered}
\end{equation}

where $\alpha$ is a hyperparameter. The ELU function is continuous and differentiable everywhere, which makes it suitable for backpropagation and gradient-based optimization.

\subsection{Swish}

Swish is an activation function proposed by Ramachandran et al. from Google Brain in 2017 \cite{ramachandran2017searching}. It is a smooth function that takes a real-valued input and produces an output between 0 and 1, making it useful for classification tasks. The Swish activation function has a learnable parameter, which allows for more efficient training of deep neural networks compared to other activation functions such as ReLU or ELU.

The Swish is formulated as:

\begin{equation}
\begin{gathered}
f(x) = x*\sigma(\beta*x)
\end{gathered}
\end{equation}

where $\sigma$ is the sigmoid function, and $\beta$ is a learnable parameter. The Swish function is also continuous and differentiable everywhere, making it suitable for gradient-based optimization. Like ELU, Swish can help mitigate the vanishing gradient problem associated with other activation functions, making it suitable for use in deeper neural networks.

\subsection{Randomized Rectified Linear Unit}
The Randomized Rectified Linear Unit (RReLU) is an activation function used in deep neural networks. It takes a real-valued input and produces an output between 0 and 1, making it useful for classification tasks. Unlike the standard ReLU function, which can suffer from dead neurons, RReLU introduces randomization to the ReLU function, resulting in a more flexible activation function with better generalization performance \cite{xu2015empirical}. The RReLU has two learnable parameters, $\alpha$ and $\beta$, that are sampled from a uniform distribution to generate a lower and upper bound, respectively. This introduces a random component to the function and allows for more efficient training of deep neural networks compared to other activation functions such as ReLU or ELU. The RReLU also helps reduce the vanishing gradient problem associated with different activation functions such as sigmoid or tanh, making it suitable for use in deeper networks where gradients can become very small over multiple layers.
\begin{equation}
\begin{gathered}
f(x) = \begin{cases}
x & \text{if } x \geq 0 \\
\alpha*x & \text{if } x < 0
\end{cases}
\end{gathered}
\end{equation}

\subsection{Randomized Rectified Linear Unit}
The Randomized Rectified Linear Unit (RReLU) is an activation function used in deep neural networks. It takes a real-valued input and produces an output between 0 and 1, making it useful for classification tasks. Unlike the standard ReLU function, which can suffer from dead neurons, RReLU introduces randomization to the ReLU function, resulting in a more flexible activation function with better generalization performance \cite{xu2015empirical}. The RReLU has two learnable parameters, $\alpha$ and $\beta$, that are sampled from a uniform distribution to generate a lower and upper bound, respectively. This introduces a random component to the function and allows for more efficient training of deep neural networks compared to other activation functions such as ReLU or ELU. The RReLU also helps reduce the vanishing gradient problem associated with different activation functions such as sigmoid or tanh, making it suitable for use in deeper networks where gradients can become very small over multiple layers.
\begin{equation}
\begin{gathered}
f(x) = \begin{cases}
x & \text{if } x \geq 0 \\
\alpha*x & \text{if } x < 0
\end{cases}
\end{gathered}
\end{equation}

\subsection{Mish}
The Mish activation function is a novel activation function that combines elements from both the ReLU and tanh functions to create a more robust non-linearity than either one alone \cite{misra2019mish}. It is defined as the product of the input and the hyperbolic tangent of the natural logarithm of 1 plus the exponential function of the input. The Mish function is smooth, monotonically increasing, and differentiable for all values of the input, allowing for efficient optimization during backpropagation. Also, it is not susceptible to the vanishing gradient problem, which makes it suitable for use in deeper networks where gradients can become very small over multiple layers \cite{weber2021riann,noel2021biologically,dubey2022activation}.

\begin{equation}
\begin{gathered}
f(x) = x*tanh(ln(1+e^x))
\end{gathered}
\end{equation}

\begin{figure}[!ht]
	\centering
	\includegraphics[width=0.5\linewidth]{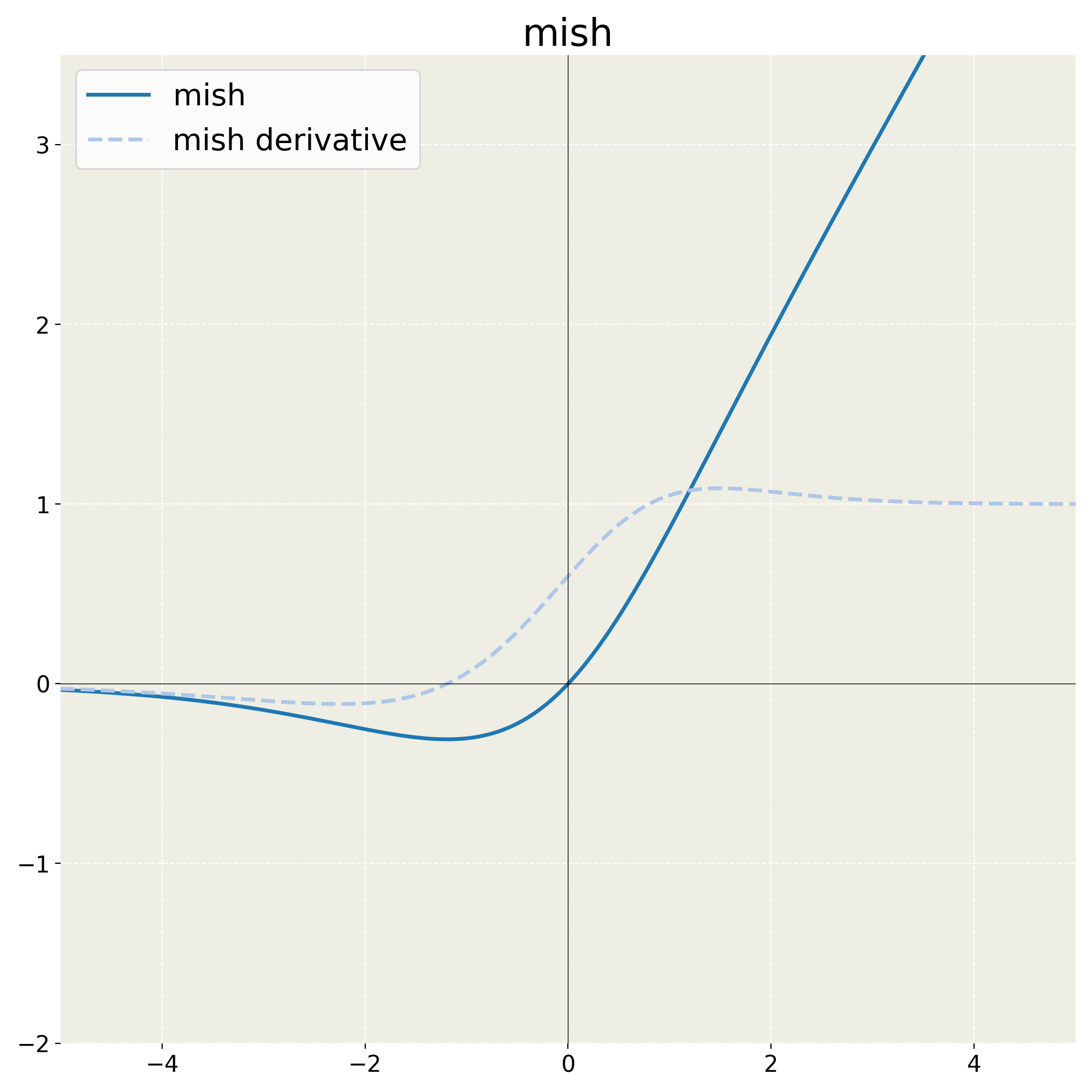}
	\caption{Mish Activation Function}
	\label{mish}
\end{figure}

Figure \ref{mish} shows the plot of the Mish activation function. The output ranges from negative infinity to positive infinity, with values close to zero representing low probabilities and values closer to infinity representing high probabilities.

\bibliographystyle{unsrt}  
\bibliography{references}

\end{document}